\documentclass[runningheads]{llncs}

\usepackage{eccv}

\usepackage{eccvabbrv}
\usepackage{tikz}
\usepackage{comment}
\usepackage{graphicx}
\usepackage{booktabs}
\usepackage{ulem}
\usepackage{xcolor}
\usepackage{color}
\usepackage{colortbl}
\usepackage{amssymb}
\usepackage{amsfonts}
\usepackage{multirow}
\usepackage{tablefootnote}
\usepackage{threeparttable}
\usepackage{booktabs}
\usepackage{pifont}% http://ctan.org/pkg/pifont
\usepackage{wrapfig}
\usepackage{tcolorbox}
\usepackage{enumitem}
\usepackage{arydshln}
\usepackage[symbol]{footmisc}
\newcommand{\cmark}{\ding{51}}%
\newcommand{\xmark}{\ding{55}}%
\definecolor{cyan}{cmyk}{.3,0,0,0}
\definecolor{mygray}{rgb}{.95,.95,.95}
\newcommand{\MODELNAME}{LHRS-Bot}
\newcommand{\ALIGNMENTNAME}{LHRS-Align}
\newcommand{\SFTNAME}{LHRS-Instruct}
\newcommand{\BENCHMARKNAME}{LHRS-Bench}
\newcommand{\VarSty}[1]{\textnormal{\ttfamily\color{cyan!90!black}#1}\unskip}

\usepackage[accsupp]{axessibility}  % Improves PDF readability for those with disabilities.
\usepackage{listings}

\usepackage{hyperref}

\usepackage{orcidlink}

\begin{document}

\title{
\MODELNAME: Empowering Remote Sensing with VGI-Enhanced Large Multimodal Language Model 
} 

\titlerunning{\MODELNAME}

\author{Dilxat Muhtar$^{\dagger}$\and
Zhenshi Li$^{\dagger}$ \and
Feng Gu \and
Xueliang Zhang* \and 
Pengfeng Xiao
}

\authorrunning{D.~Muhtar et al.}

\institute{Nanjing University \\
\email{Lzhenshi@outlook.com, \{dmuhtar, gufeng\}@smail.nju.edu.com \\ \{zxl, xiaopf\}@nju.edu.cn}
}

\maketitle

\renewcommand{\thefootnote}{\fnsymbol{footnote}}
\setcounter{footnote}{0} % Start with dagger symbol
\footnotetext[2]{Equal contribution, listed in random order.}
\footnotetext[1]{Corresponding author.}
\renewcommand{\thefootnote}{\arabic{footnote}}
\setcounter{footnote}{0} 

\begin{abstract}
The revolutionary capabilities of large language models (LLMs) have paved the way for multimodal large language models (MLLMs) and fostered diverse applications across various specialized domains.
In the remote sensing (RS) field, however, the diverse geographical landscapes and varied objects in RS imagery are not adequately considered in recent MLLM endeavors. 
To bridge this gap, we construct a large-scale RS image-text dataset, LHRS\footnote{LHRS stands for 'Language Helps Remote Sensing'.}-Align, and an informative RS-specific instruction dataset, \SFTNAME, leveraging the extensive volunteered geographic information (VGI) and globally available RS images.
Building on this foundation, we introduce \MODELNAME, an MLLM tailored for RS image understanding through a novel multi-level vision-language alignment strategy and a curriculum learning method.
Additionally, we introduce \BENCHMARKNAME, a benchmark for thoroughly evaluating MLLMs' abilities in RS image understanding.
Comprehensive experiments demonstrate that \MODELNAME~exhibits a profound understanding of RS images and the ability to perform nuanced reasoning within the RS domain\footnote{Data, Code and model are available at \url{https://github.com/NJU-LHRS/LHRS-Bot}}.

\end{abstract}

\section{Introduction}
\label{sec:intro}

The recently advanced large language models (LLMs) have presented remarkable capabilities in engaging in informative conversations and solving complex problems~\cite{touvron2023llama,touvron2023llama2,chowdhery2023palm,brown2020GPT3,jiang2024mixtral,MosaicML2023Introducing,bai2023qwenreport}. 
To extend the perceptual capabilities of LLMs, multimodal large language models (MLLMs) bring eyes to LLMs. These models, by aligning with visual representations and visual instruction tuning, have demonstrated impressive multimodal instruction following capabilities and can act as
a general-purpose interface for a broad range of tasks~\cite{li2023blip,alayrac2022flamingo,huang2023kosmos,liu2023visual, bai2023qwen,zhu2023minigpt,li2023multimodal}. 
\begin{wrapfigure}{r}{0.55\textwidth}
    \begin{minipage}{0.55\textwidth}
        \centering  
        \vspace{-2mm}
        \scalebox{1.00}
        {
            \includegraphics[width=\textwidth]{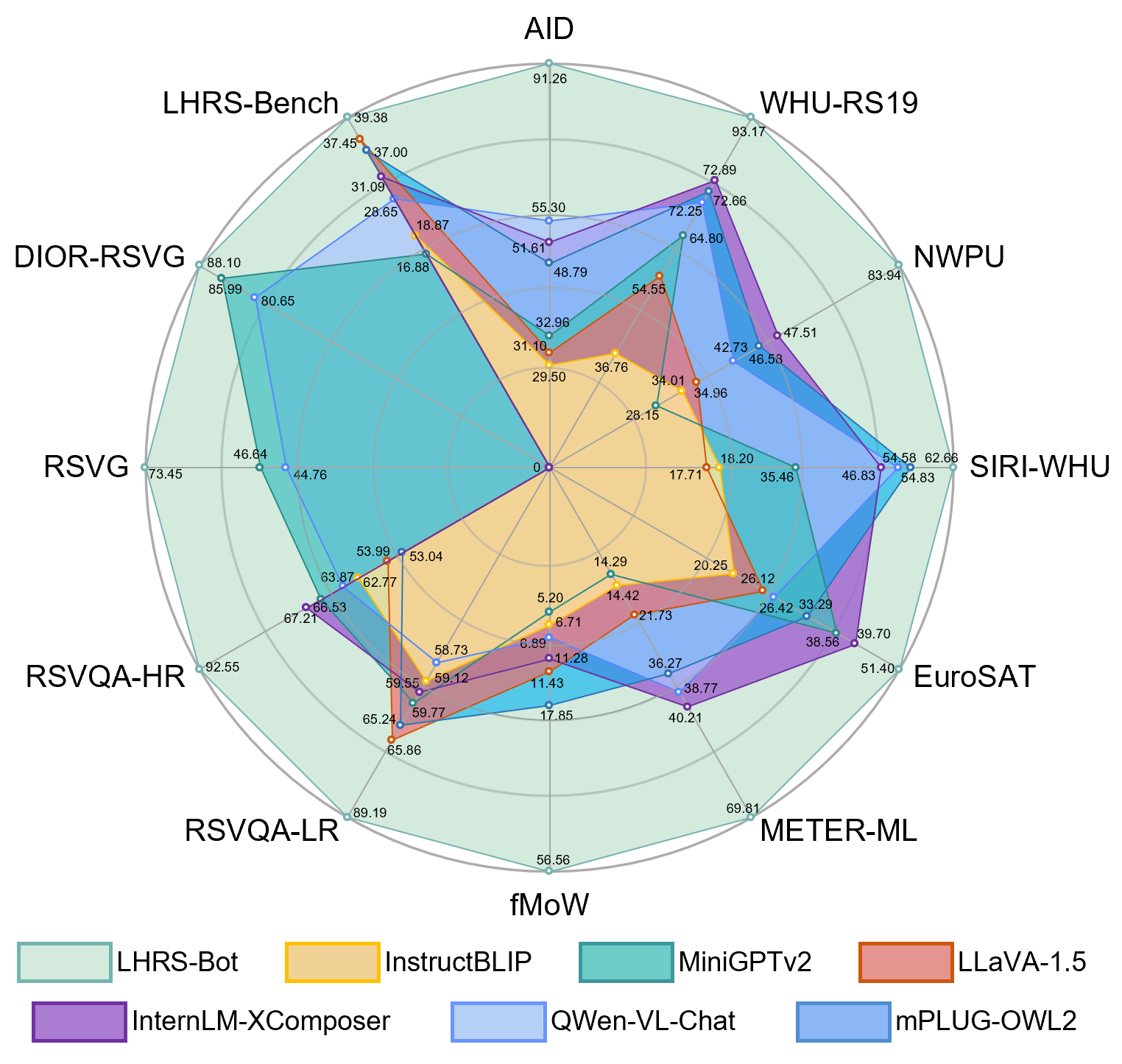}  
        }
        \captionof{figure}{Overall performance comparison between \MODELNAME~and existing MLLMs on different RS image understanding tasks.}
        \label{fig:radar}
        \vspace{-8mm}
    \end{minipage}
\end{wrapfigure}
This success in general domain has served as a catalyst for the development of domain specialized MLLMs, with notable applications in biomedicine~\cite{li2023llava}, autonomous driving~\cite{ma2023dolphins}, and robotics~\cite{driess2023palm}.

Among various specialized domains, remote sensing (RS) image understanding is particularly important, as it directly involves the monitoring and analysis of the Earth's surface and environment~\cite{yang2013role,ratledge2022using,wen2023survey,bashmal2023survey}.
However, RS images present significant challenges for holistic image understanding due to the intricate complexity of different objects across diverse landscapes worldwide, coupled with the variability in visual scales~\cite{muhtar2023cmid,hossain2019segmentation,reed2023scale}.
MLLMs bring together vision and language,
offering possibilities for a unified and interactive understanding of RS images~\cite{wen2023survey}.
In this direction, several RS-specific MLLMs have been proposed, such as RSGPT~\cite{hu2023rsgpt}, GeoChat~\cite{kuckreja2023geochat}, and SkyEyeGPT~\cite{zhan2024skyeyegpt}.
However, they are struggling to fit existing public RS datasets, due to the oversight of exploiting the extensive and worldwide RS features to infuse comprehensive RS visual knowledge into LLMs. 
Furthermore, these contemporary approaches of bridging vision and language domains tend to focus on the incorporation of high-level visual semantics. 
This strategy neglects the fact that different levels of visual information are crucial to fully align the linguistic and visual domains, which could facilitate a unified RS image understanding at different granularity~\cite{muhtar2023cmid,xia2018dota,wang2023samrs}. 
Detailed related works are available in supplementary material.

To bridge the gap, we present \MODELNAME, a specialized MLLM for RS, enhanced by globally available volunteered geographical information (VGI) and worldwide RS images.
Specifically, we construct a large-scale, semantically rich, and feature-diverse dataset, \ALIGNMENTNAME, by geographically pairing RS images 
% from Google Earth\footnote{\url{https://www.google.com/earth/}} 
with plentiful attributed information from the OpenStreetMap VGI database\footnote{\url{https://www.openstreetmap.org/}}, and consequently generating image captions using LLMs.
Our rigorous data cleaning process, which includes deduplication, pruning, and semantic balancing, ensures that \ALIGNMENTNAME~comprises 1.15 million meaningful and high-quality RS image-text pairs, laying a solid foundation for embedding rich RS visual knowledge into MLLMs.
Furthermore, we construct our instruction dataset, \SFTNAME, by reorganizing a broad range of open-source RS datasets into multi-task instructional datasets and employing GPT-4~\cite{openai2023gpt4} to create complex instructional data from the high-quality subset of \ALIGNMENTNAME.
Building on these datasets, we propose a novel RS-specific MLLM, \MODELNAME, by implementing a novel bridging strategy for efficiently summarizing multi-level visual representations, alongside a curriculum learning methodology. \MODELNAME~achieves \textbf{state-of-the-art} performance across a variety of RS image understanding tasks (\cref{fig:radar}) and demonstrates remarkable abilities to detect intricate objects, engage with human conversation, and provide insights from visual information within RS images (\cref{fig:qualitative_exp}).

Finally, we manually construct a high-quality benchmark, \BENCHMARKNAME, to facilitate the RS community in evaluating the RS-domain MLLMs across diverse evaluation dimensions. 
\BENCHMARKNAME~includes 690 single-choice questions, spanning 5 top-level evaluation dimensions including 11 fine-grained categories to facilitate a comprehensive, objective, and quantitative RS-specific evaluation.

The main contributions of our work are summarized as follows: 
\begin{itemize}
    \item[1.] We introduce \textbf{\ALIGNMENTNAME}, a large-scale RS image-text dataset, created through a meticulous data generation pipeline using open-source global geographic data. \ALIGNMENTNAME~showcases a broad spectrum of RS semantic visual information, establishing a solid foundation for RS-specific MLLMs.
    \item[2.] We create \textbf{\SFTNAME}, a multimodal instruction-following dataset tailored for RS image understanding. \SFTNAME~not only contains instruction data to serve for various RS image understanding tasks but also contains complex visual reasoning data generated by GPT-4.
    \item[3.] We propose \textbf{\MODELNAME}, an MLLM specifically designed for the RS domain. \MODELNAME~employs a novel bridging strategy and leverages a curriculum learning approach to fully exploit the inherent knowledge of the proposed datasets. \MODELNAME~demonstrates superior performance across various RS image understanding tasks, along with the ability to follow complex instructions.
    \item[4.] We establish \textbf{\BENCHMARKNAME}, a benchmark aimed at providing a comprehensive and systematic evaluation of RS MLLMs. \BENCHMARKNAME~comprises 690 meticulously crafted single-choice questions, encompassing a broad spectrum of 11 fine-grained dimensions.
\end{itemize}

\section{Dataset}
\label{sec:dataset}
\subsection{Large-Scale RS Alignment Dataset——\ALIGNMENTNAME}
\label{sec:dataset:alignment}

\begin{figure}[t]
  \centering
  \includegraphics[width=\linewidth]{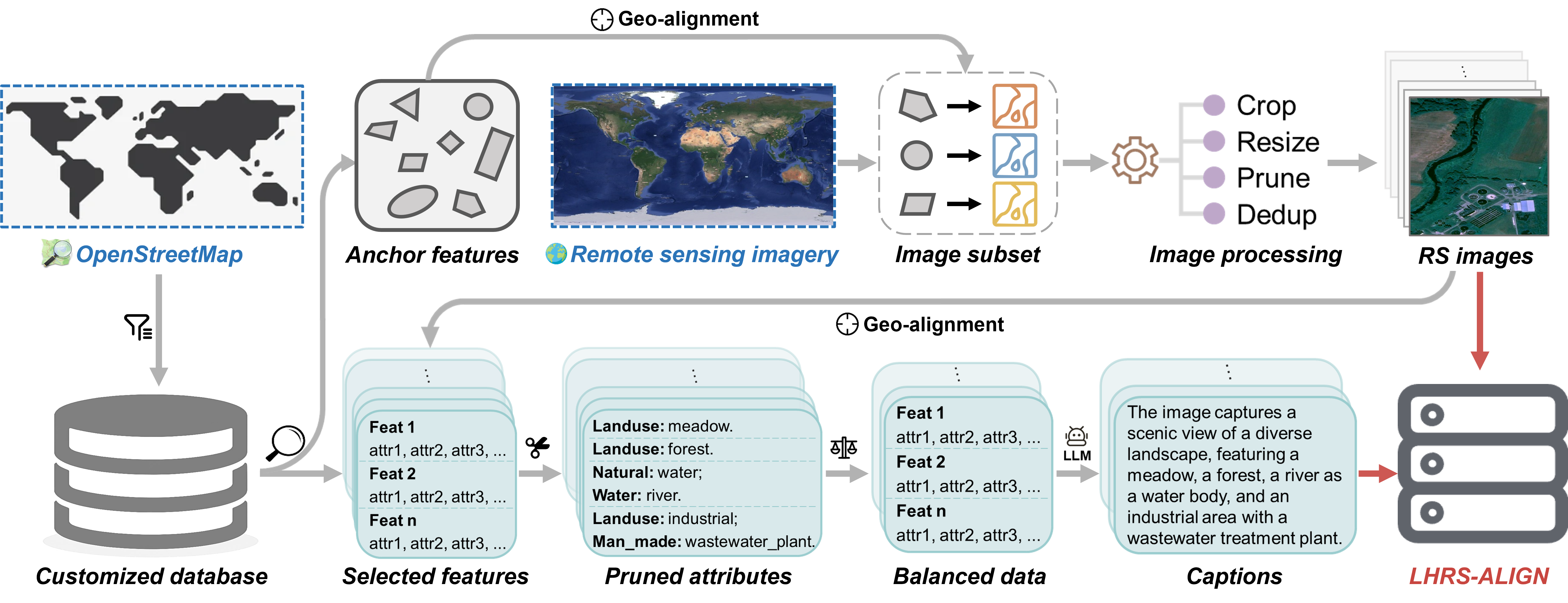}
  \caption{Pipeline of constructing \ALIGNMENTNAME.}
  \label{fig:pipeline}
  \vspace{-0.5cm}
\end{figure}

\ALIGNMENTNAME~is developed by initially geo-aligning orthorectified RS images with geographic vector features from the free spatial database, OSM. The attributes of these features are then utilized to generate image captions employing an LLM. The detailed pipeline is illustrated in \cref{fig:pipeline} and introduced as below.

\textbf{Data Preparation.} 
The RS images of \ALIGNMENTNAME~are sourced from Google Earth imagery\footnote{\url{https://www.google.com/earth/}}, encompassing a worldwide coverage with a spatial resolution of 1 meter. It should be noted that our proposed pipeline is applicable to any RS imagery with geographic coordinates. The geographic features are gathered from OSM, a crowdsource VGI database containing abundant geographical objects. Each object (feature) in the OSM database is annotated as the digitization of the object geometry (e.g., polygons, lines, and points) as well as the semantic tags. A tag is a key-value pair that provides information about a geographic object\footnote{\url{https://wiki.openstreetmap.org/wiki/Map\_features}}, for example, “landuse=residential”. 
Thanks to the free access to the globally encompassing data sources, our pipeline can produce a dataset that offers a vast and comprehensive variety on a worldwide scale.
For the current version of \ALIGNMENTNAME, we collect RS images and OSM features within a scope of 0.3° longitude × 0.28° latitude for each of the selected cities from all over the world. 
We customize our own OSM database to streamline the subsequent feature selection process. More information about our data source and database is introduced in supplementary material.

\textbf{Geo-alignment between Images and Geographic Features.} We iterate through our customized database to identify qualified features serving as “anchors”. Based on the geographical extent of each anchor feature, the corresponding RS image is acquired, which is a square bounding box centered on the feature. These images subsequently undergo a stream of postprocessing and pruning procedures. Following this, due to the possibility of multiple features being present within a single image extent, all the features within each image are further identified. At this point, we have established a raw data pool of around 4 million pairs, in which one RS image corresponds to multiple geographical features. Further details for this processes are provided in supplementary material.

\textbf{Attribute Pruning and Semantic Balancing.} In the OSM database, all the features are assigned with their semantic attributes, that is, multiple key-value tags. However, not all the keys are favorable information for visual identification, such as website and postal code. Hence, we filter out irrelevant keys through a pipeline as detailed in supplementary material. \cref{fig:pipeline} provides an example showcasing an image along with its geographical features and their corresponding attributes. 
Furthermore, the OSM database exhibits considerable feature imbalance, notably dominated by geographical features such as residential or farmland. 
Given the importance of balanced semantic information for effective vision-language alignment~\cite{radford2021learning, xu2023demystifying}, we implement a data balancing method, as detailed in supplementary material, to create a more semantically balanced dataset. 
As a result, this balancing process retains 1.15 million images remaining out of the original 4 million. 

\textbf{Caption Generation with LLM.} 
The use of image-caption pairs to incorporate visual knowledge into LLMs has been validated as an effective approach~\cite{li2023llava,zhu2023minigpt,ye2023mplug,dai2305instructblip,bai2023qwen}.
However, using plain image key-value labels can be problematic, as they lack grammatical structure, potentially disrupting the generative and comprehension abilities of LLMs.
Similarly, rule-based caption generation strategies may introduce significant redundancy.
Therefore, we opt to use open-source LLMs to generate image captions by summarising the key-value tags of the corresponding images.
We experiment with various in-context learning examples and LLMs, ranging in size from 7B to 70B. 
After comprehensive testing by a domain expert, we select Vicuna-v1.5-13B~\cite{chiang2023vicuna}~for caption generation as it achieves the best trade-offs in terms of generation efficiency and quality. 
Detailed information about the in-context examples and the specific deployment of Vicuna-v1.5-13B~\cite{chiang2023vicuna}~can be found in supplementary material.
Our final dataset, referred as \ALIGNMENTNAME, comprises approximately 1.15 million image-caption pairs.
Each pair is accompanied with additional metadata, such as image resolution and geographical location (country and city).
Several examples of \ALIGNMENTNAME~are also presented in supplementary material.

\subsection{RS-Specific Instruction-Following Dataset——\SFTNAME}
We craft a multimodal instruction-following dataset~\SFTNAME~by utilizing two public image-caption datasets and our \ALIGNMENTNAME~dataset.

\textbf{Instruction-Following Dataset from Public RS Caption Dataset.}
We start by meticulously evaluating and selecting all image-caption datasets within the RS domain. 
Our selection process focuses on the quality of captions and involves a thorough human evaluation. 
This requires each image to be accompanied by five distinct captions, with each caption uniquely detailing different aspects of the image. Based on this criterion, we identify RSITMD~\cite{yuan2022exploring} and NWPU~\cite{cheng2017remote} as the datasets with superior caption quality.
Subsequent to selection, we design a rigorous data cleaning process, which involves deduplication, computation of caption lengths, and assessment of relevance through CLIP-score calculations. 
This results in the extraction of 3.4K and 30K high-quality image-caption pairs from RSITMD and NWPU, respectively.
We then use several in-context examples along with the five captions per image for prompting Vicuna-v1.5-13B~\cite{chiang2023vicuna} to generate a conversation that specifically focuses on the content of the objects in the image.
Finally, we filter out the most relevant conversations through pre-defined rules and evaluations from three domain experts. 
This process results in 2.9K and 25K instruction data from these two datasets, respectively.
For more detailed information on the in-context examples used and the data filtering criteria, please refer to the supplementary material.

\textbf{Instruction-Following Dataset from~\ALIGNMENTNAME.}
\label{sec:dataset:sft}
Considering the limitations of public image-caption datasets, which typically lack bounding box annotations, and the constrained capabilities of Vicuna-v1.5-13B~\cite{chiang2023vicuna}, the generated conversations might omit crucial details such as the locations and quantity of objects in the images, along with visual reasoning information.
To mitigate this issue, we meticulously select 15K representative RS images that are rich in attribute information from our~\ALIGNMENTNAME~dataset.
We then proceed to calculate the bounding boxes for various objects within these images, utilizing the built-in spatial coordinates of different attributes in the OSM database. 
Leveraging this enriched data, we prompt GPT-4 to craft complex instruction data that incorporates visual reasoning, detailed image descriptions, and conversations about the location and number of objects.
The final step involves a manual cleaning of the generated results. This process yields a dataset comprising 0.9K samples about visual reasoning, 4K samples about detailed image description, and 7K samples about conversation.
For more details, please refer to supplementary material.
% \vspace{-0.3cm}

\subsection{RS-specific MLLM Evaluation Benchmark——\BENCHMARKNAME}
We construct \BENCHMARKNAME~to evaluate the performance of MLLMs in the RS domain.
\BENCHMARKNAME~employs hierarchical taxonomies to comprehensively evaluate MLLMs across various dimensions (\cref{tab:benchmark}). 
To prevent data leakage, we abstain from using public RS datasets and instead meticulously collect images from Google Earth.In terms of question format, we utilize single-choice questions
\begin{wrapfigure}{r}{0.5\textwidth}
    \begin{minipage}{0.5\textwidth}
        \centering  
        \vspace{-8mm}
        \captionof{table}{Overview of the dataset distribution. \BENCHMARKNAME~contains 108 RS images with 690 question-answer pairs, covering five major evaluation dimensions and 11 sub-dimensions of questions. Each sample is a human-annotated single-choice question that may involve one or more evaluation dimensions simultaneously.}
        \centering
        \label{tab:benchmark}
        \scalebox{0.88}
        {
            \begin{tabular}{lcc}
            \toprule \rowcolor{mygray}
            Dimension& Sub-dimension & Count    \\ \midrule
            \multirow{5}{*}{Recognition} & Identity & 634       \\
                                        & Color & 113              \\
                                        & Orientation & 39        \\
                                        & Shape & 37              \\
                                        & Area & 75               \\ \hline
            \multirow{2}{*}{Imagery}    & Resolution & 21         \\ 
                                        & Modality & 23           \\ \hline
            \multirow{2}{*}{Spatial-awareness} & Location & 204    \\ 
                                        & Distance & 22          \\ \hline
            Quantity                   & - & 137                  \\ \hline     
            Reasoning                   & - & 46          \\      \bottomrule
            \end{tabular}
        }
        \vspace{-8mm}
    \end{minipage}
\end{wrapfigure}
~\cite{fu2023mme, liu2023mmbench, li2023seed1, li2023seed2} with 2 to 4 candidate answers, enabling a more quantitative and objective evaluation compared to open-ended questions~\cite{bai2023touchstone,yu2023mmvet,bitton2023visit, liu2023visual}. Every question and answer within \BENCHMARKNAME~are manually crafted to guarantee precision and reliability. Totally, \BENCHMARKNAME~comprises 108 images and 690 questions. Some examples of \BENCHMARKNAME~are shown in supplementary material.

The evaluation dimensions of \BENCHMARKNAME~encompass 5 top-level dimensions, further delineated into 11 fine-grained sub-dimensions, as shown in \cref{tab:benchmark}. 
These top-level dimensions include recognition, imagery, spatial-awareness, quantity, and reasoning. 
For detailed explanations about each dimensions, please refer to supplementary material.

\section{Method}
\label{sec:method}
\begin{figure}[!t]
  \centering
  \includegraphics[width=\linewidth]{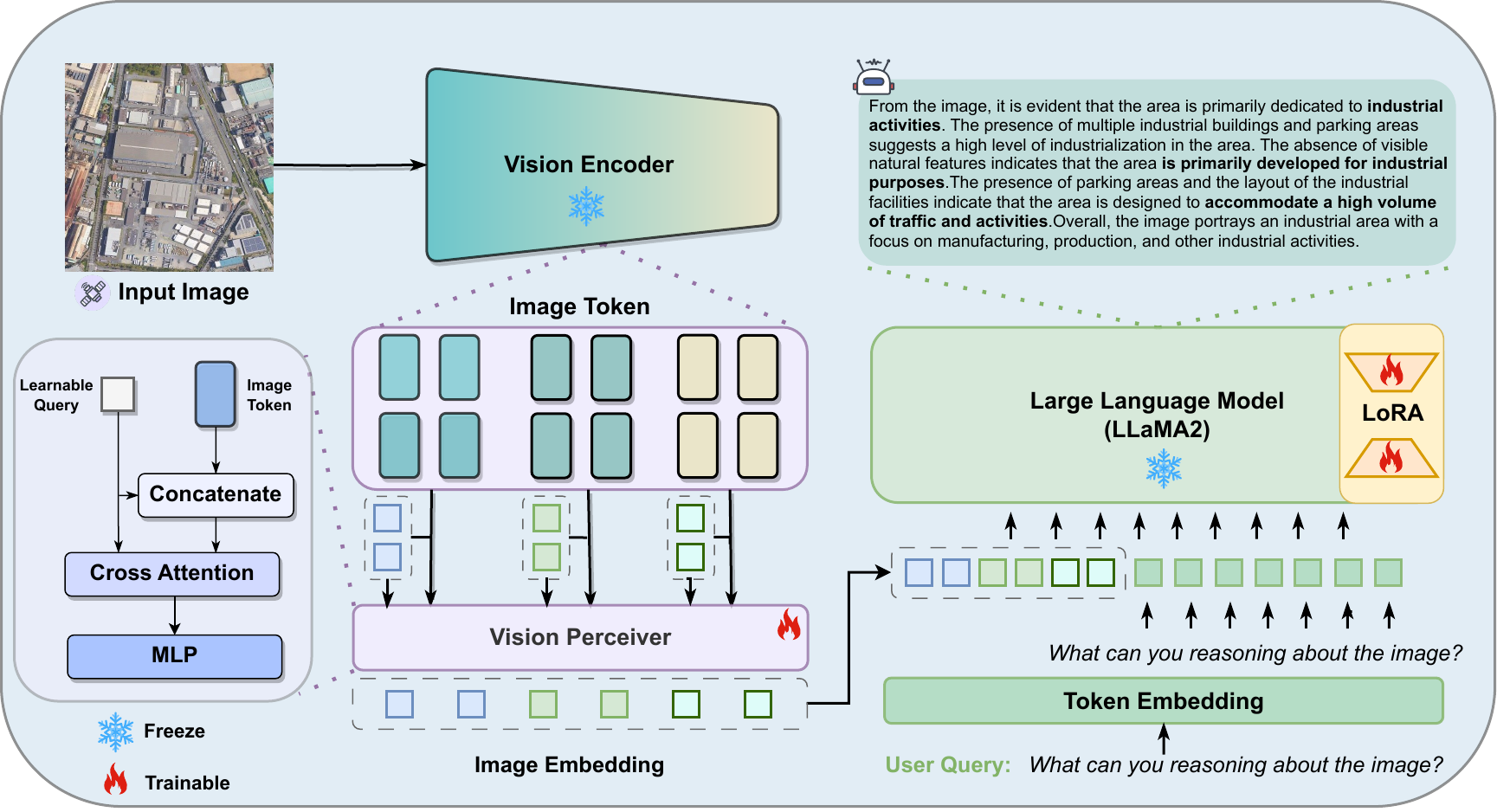}
  \caption{Architecture of the proposed \MODELNAME. \MODELNAME~employs learnable queries in conjunction with a vision perceiver to summarize multi-level visual representations. These representations are subsequently concatenated with language token embeddings for processing within the LLM.}
  \label{fig:architecture}
  \vspace{-5mm}
\end{figure}
We present \MODELNAME~to harness the full potential of LLMs and general visual encoders for enhanced RS images understanding and reasoning.
This section delves into each components of \MODELNAME, detailing how we achieve fully alignment between vision and language models across multiple levels. 
Subsequently, we explain the curriculum training strategy for training \MODELNAME.

\subsection{Model Architecture}
The overall architecture of \MODELNAME~is presented in~\cref{fig:architecture}. 
\MODELNAME~is primarily composed of three components: a universal vision encoder, a vision perceiver, and a foundational LLM.

\textbf{Vision Encoder.} 
\MODELNAME~utilizes the pre-trained CLIP visual encoder ViT-L/14~\cite{radford2021learning} with input resolution of 224$\times$224 pixel for encoding images. 
The prior works~\cite{zhu2023minigpt,liu2023visual,ye2023mplug} usually utilize the final hidden feature for abstracting the semantic content of an image.
However, several studies have highlighted that features at different levels of a deep learning model embed distinct types of information~\cite{ghiasi2022vision,li2021visualizing,park2023self}.
Therefore, we argue that previous image encoding methods fail to sufficiently capture the semantic content of images, leading to a reliance on extensive image-text pairs for vision-language alignment. 
Consequently, we retain multiple hidden features from various layers of the image encoder. This approach offers a more thorough representation of the underlying image, facilitating a thorough and efficient domain alignment.

\textbf{Vision Perceiver.}
Incorporating multi-level visual features brings extended image tokens, increasing computational overhead and potentially overwhelming language information. 
To mitigate this issue, we introduce a set of learnable queries for each level of image features. 
These queries are designed to summarize the semantic information of each level of visual features through a series of stacked cross-attention and MLP layers. 
Once summarized, these image features are then fed into the LLM.
Furthermore, we implement a descending query allocation strategy inspired by the observation that image feature redundancy tends to increase with network depth~\cite{ju2023turbo,bolya2022token}. 
This strategy allocates more queries to the shallower image feature levels while gradually reducing the number of queries for deeper levels.

\textbf{Large Language Model.}
We employ the LLaMA2-7B~\cite{touvron2023llama2} as the central expert for interpreting visual and language information to output the desired results following the previous practives~\cite{saygin2023quilt,li2023llava,kuckreja2023geochat}.
\begin{figure}[t]
  \centering
  \includegraphics[width=\linewidth]{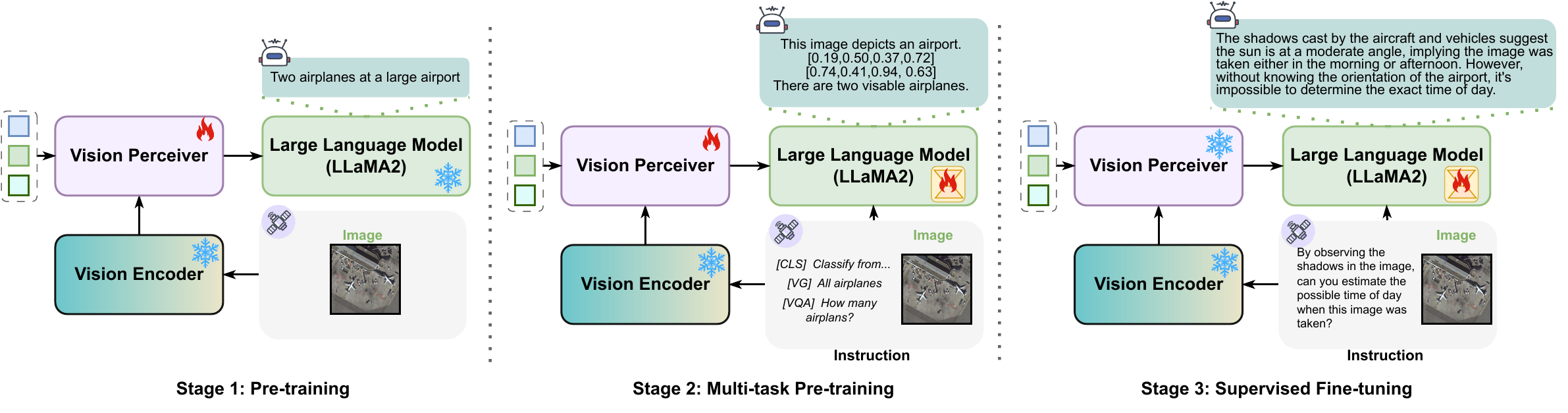}
  \caption{Curriculum learning strategy. We progressively unlock various parameters and introduce data of escalating complexity at different stages, enabling the model to gradually assimilate visual knowledge with increasing difficulty.}
  \label{fig:learning}
  \vspace{-0.5cm}
\end{figure}
\vspace{-0.3cm}
\subsection{Curriculum Learning Strategy}
We implement a three-stage curriculum learning strategy by progressively increasing the difficulty of training tasks~(\cref{fig:learning}). 
This strategy incrementally aligns the vision encoder with the LLM, ultimately enabling the LLM to solve multiple visual tasks and follow complex vision-language instructions.

\textbf{Stage~1:~Pre-training.} 
In the first stage, we use the large-scale, weakly labelled~\ALIGNMENTNAME~dataset (\cref{sec:dataset:alignment}) to train the vision perceiver to incorporate a wide range of RS visual knowledge into the LLM.
Specifically, for each sample in the~\ALIGNMENTNAME~dataset, the output visual embedding tokens $X_v$ obtained from the vision perceiver are concatenated with the language embedding tokens $X_l$ from the corresponding image caption. 
The concatenated result is then fed into the LLM. 
The objective of this stage is to maximise the conditional probability $P(X_l|X_v;\theta)$ of generating the caption tokens $X_l$ given the prefix visual tokens $X_v$:
\begin{align}
\label{eq:stage1obj}
    \begin{split}
      \mathcal{L} & = \log P(X_l|X_v;\theta) \\
                  & = \sum_{i=1}^L\log P(x_i|X_v, X_{l,<i};\theta),
    \end{split}
    \vspace{-5mm}
\end{align}
where $L$ is the length of caption tokens, $\theta$ represents the trainable parameters, and $X_{l,<i}$ denotes the generated tokens prior to the current prediction token $x_i$.

\textbf{Stage~2: Multi-task Pre-training.}
In the second stage, we further train the vision perceiver and efficiently fine-tune the LLM using LoRA~\cite{hu2021lora} on multi-task instruction datasets to improve multimodal task solving capabilities of LLM.
The construction of the multi-task instruction datasets begins with the collection of diverse public datasets in the RS domain, encompassing tasks such as classification, VQA, and visual grounding. 
These collected datasets are then meticulously transformed into instruction-based formats using manually designed templates.
More detailed information about these multi-task datasets and the templates used can be found in the supplementary material.
Moreover, we integrate the instruction data from NWPU, RSITMD and the 4K detailed descriptions data from~\SFTNAME~(\cref{sec:dataset:sft}) to further enrich the LLM with fine-grained RS visual knowledge.
The objective for this stage remains similar to~\cref{eq:stage1obj} from stage 1, albeit with an adaptation that the generated tokens are now additionally conditioned on the instruction tokens $X_{instruct}$:
\begin{align}
    \begin{split}
      \mathcal{L} & = \log P(X_l|X_v, X_{instruct};\theta) \\
                  & = \sum_{i=1}^L\log P(x_i|X_v, X_{instruct}, X_{l,<i};\theta).
    \end{split}
\end{align}

\textbf{Stage~3: Supervised Fine-tuning.}
In the final stage, our focus shifts towards enhancing the conversational and reasoning capabilities of our model by incorporating more complex instruction data. 
We utilize all the instruction data from the~\SFTNAME~dataset (\cref{sec:dataset:sft}) and randomly select 20K data from the LLaVA complex reasoning dataset~\cite{li2023llava}. 
Additionally, we mix the above datasets with the multi-task datasets from stage 2 using a lower data sampling ratio to further unleash the multi-task problem-solving ability of~\MODELNAME.
\vspace{-3mm}

\section{Experiment}
\label{sec:experiment}
We conduct comprehensive experiments across image classification, VQA, and visual grounding tasks within the RS domain. 
These experiments are designed to rigorously validate the superior multi-task solving capabilities of \MODELNAME. 
Furthermore, we evaluate the performance of various LLMs in the RS domain using the proposed \BENCHMARKNAME~benchmark. 
We present both quantitive and quality results.
\subsection{Experimental Setup}
\label{section:experi_setup}
\subsubsection{Dataset.}
We select diverse RS datasets for providing a comprehensive assessment across different tasks within the RS field: seven RS classification datasets (namely AID~\cite{xia2017aid}, WHU-RS19~\cite{Dai2011WHURS19}, NWPU~\cite{cheng2017remote}, SIRI-WHU~\cite{zhu2016SIRI-WHU}, EuroSAT~\cite{helber2019eurosat}, METER-ML~\cite{zhu2022meter}, and fMoW~\cite{christie2018functional}), two VQA datasets (LR and HR subsets of RSVQA~\cite{lobry2020rsvqa}), and two visual grounding datasets (RSVG~\cite{sun2022VG} and DIOR-RSVG~\cite{rsvg}). 
We exclude any data from the test sets that overlap with the training set to avoid data leakage.
More details about each test dataset are presented in supplementary material.
\vspace{-0.3cm}
\subsubsection{Baseline.}
We evaluate~\MODELNAME~and various powerful open-source MLLMs, including LLaVA-v1.5~\cite{liu2023improved}, MiniGPT-v2~\cite{zhu2023minigpt}, InstructBLIP~\cite{dai2305instructblip}, mPLUG-Owl2~\cite{ye2023mplug}, QWen-VL-Chat~\cite{bai2023qwen}, and InternLM-Xcomposer~\cite{zhang2023internlm}, on different tasks.
Furthermore, we also include the results of RSGPT~\cite{hu2023rsgpt}, GeoChat~\cite{kuckreja2023geochat} and SkyEyeGPT~\cite{zhan2024skyeyegpt} from the original papers for comparisons whenever possible.
Unless otherwise noted, we utilized 7B variants for all our experiments.

\vspace{-0.3cm}
\subsubsection{Implementation.}
We extract hidden states at layers $\{i=L/3, j=2L/3, k=L - 1\}$ for summarizing through vision perceiver, where $L$ denotes the number of layers in the vision encoder.
Our vision perceiver implementation comprises six layers of cross-attention and MLP, with each hidden state assigned $\{64, 48, 32\}$ learnable queries.
We apply LoRA module to every linear layer of the LLM, setting the rank $r$ and the scale factor $\alpha$ for LoRA to $128$ and $256$, respectively.
Moreover, we add task identifier $\text{[CLS]}, \text{[VQA]}$, and $\text{[VG]}$ for classification, VQA, and visual grounding tasks, respectively, for enabling~\MODELNAME~to better distinguish different tasks~\cite{zhu2023minigpt}. 
For additional details on the hyperparameters used at each stage, please refer to supplementary material.

\begin{table}[t!]
\caption{Classification accuracy of different models on each dataset.}
\vspace{-0.2cm}
\label{tab:cls_compare}
\resizebox{\columnwidth}{!}{%
\begin{tabular}{lcccccccc}
\toprule \rowcolor{mygray}
\multicolumn{1}{c}{Method} & AID   & WHU-RS19 & NWPU         & SIRI-WHU & EuroSAT & METER-ML & fMoW     & Avg.  \\ \midrule
LLaVA-1.5                & 31.10 & 54.55    & 34.96         & 17.71    & 26.12   & 21.73    & 11.43  & 28.23 \\
MiniGPTv2                & 32.96 & 64.80    & 28.15         & 35.46    & 38.56   & 14.29    & 5.20   & 31.35 \\ 
InstructBLIP             & 29.50 & 36.76    & 34.01         & 18.20    & 20.25   & 14.42    & 6.71   & 22.84 \\
mPLUG-OWL2               & 48.79 & 72.66    & 46.58         & \underline{54.83}    & 33.29   & 36.27    & \underline{17.85}  & \underline{44.32} \\
QWen-VL-Chat             & \underline{55.30} & 72.25    & 42.73         & 54.58    & 26.42   & 38.77    & 6.89   & 42.42 \\
InternLM-XComposer        & 51.61 & \underline{72.89}    & \underline{47.51}         & 46.83    & \underline{39.70}   & \underline{40.21}    & 11.28  & 44,29 \\ \rowcolor{cyan!50}
\MODELNAME              & \textbf{91.26 }& \textbf{93.17}    & \textbf{83.94}         & \textbf{62.66}    & \textbf{51.40}    & \textbf{69.81}   & \textbf{56.56} & \textbf{71.83} \\ \bottomrule
\end{tabular}
}
\vspace{-0.25cm}
\end{table}
\subsection{Quantitative Evaluation}
We evaluate different MLLMs using greedy decoding with the same prompt for each task and report the average result of two independent trials. 
Details about the evaluation prompts are provided in supplementary material.
\vspace{-0.3cm}

\subsubsection{Classification.}
The classification results across different datasets are presented in~\cref{tab:cls_compare}.
These results show that~\MODELNAME~outperforms all competing models in all seven classification datasets, with a notable margin of 27.51\% higher average accuracy than the second ranked method.
It is worth mentioning that, except for NWPU, METER-ML, and fMoW, the remaining datasets are completely absent from the multi-task training dataset 
\footnote{However, we have no means of confirming whether these data appeared in the \ALIGNMENTNAME~dataset, considering that images in \ALIGNMENTNAME, along with several of the classification datasets (AID, WHU-RS19, and SIRI-WHU), are collected from Google Earth.}, demonstrating the strong generalization ability of~\MODELNAME.
In addition, in the context of the fine-grained RS classification dataset fMoW, our method significantly surpasses other models, leading by an impressive 40\% on average. 
This highlights the critical role of continue pre-training in the RS domain and the effectiveness of our alignment dataset~\ALIGNMENTNAME.
% \vspace{-0.3cm}

\begin{table}[t!]
\caption{Accuracy of different models on each visual question answering dataset. Following the practice of prior works~\cite{hu2023rsgpt,kuckreja2023geochat,zhan2024skyeyegpt}, we omit area and count questions during evaluation.}
\vspace{-0.2cm}
\label{tab:vqa_compare}
\centering
\scalebox{0.88}
{
\begin{tabular}{l|cccc|ccc}
\hline \rowcolor{mygray}
Method & \multicolumn{4}{c|}{RSVQA-LR} & \multicolumn{3}{c}{RSVQA-HR}                                                               \\
                         & Rural/Urban & Presence & Compare & Avg. & Presence                     & Compare              &   Avg.   \\ \hline
LLaVA-1.5                & 59.22       & 73.16    & 65.19   & 65.86 & 48.96                           & 59.02                &  53.99    \\
MiniGPTv2                & 60.02       & 51.64    & 67.64   & 59.77 & 68.34                           & 64.71                &  66.53    \\
InstructBLIP             & 62.62       & 48.83    & 65.92   & 59.12 & 62.63                           & 62.90                &  62.77    \\
mPLUG-Owl2               & 57.99       & 74.04    & 63.69   & 65.24 & 47.60                           & 58.47                &  53.04    \\
QWen-VL-Chat             & 62.00       & 47.65    & 66.54   & 58.73 & 61.75                           & 65.98                &  63.87    \\ 
InternLM-XCompose        & 59.00       & 66.74    & 52.91   & 59.55& 67.79                         & 66.62                   &  67.21    \\ \hline
RSGPT                    & \textbf{94.00}       & \textbf{91.17}    & \underline{91.70}  & \textbf{92.29} & \underline{90.92}          & \underline{90.02}        &  \underline{90.47}    \\
GeoChat                  & \underline{91.09}       & \underline{90.33}    & \textbf{94.00} & \underline{91.81} & 58.45                      & 83.19                 &  70.82 \\
SkyEyeGPT                & 88.93       & 88.63    & 75.00  & 84.16 & 80.00                      & 80.13                 &  82.56    \\\rowcolor{cyan!50} 
\MODELNAME               & 89.07       & 88.51    & 90.00  & 89.19 & \textbf{92.57}             & \textbf{92.53}        &  \textbf{92.55}    \\\hline
\end{tabular}
}
\vspace{-0.3cm}
\end{table}

\subsubsection{Visual Question Answering.}
We report the results of different models on RSVQA-LR and RSVQA-HR datasets in~\cref{tab:vqa_compare}.
\MODELNAME~demonstrates performance on par with the previously leading RS domain-specific models on the VQA task. 
Notably, \MODELNAME~achieves comparable results to RSGPT, despite RSGPT being fine-tuned for five epochs on the RSVQA datasets. 
Furthermore, \MODELNAME~sets a new record in the RSVQA-HR task with high-resolution RS images, showcasing its superior visual recognition abilities with high-resolution RS images.
\vspace{-0.3cm}

\begin{wrapfigure}{r}{0.5\textwidth}
    \begin{minipage}{0.5\textwidth}
        \centering  
        \vspace{-12mm}
        \captionof{table}{Comparison of visual gronding tasks. Accuracy is calculated based on whether the predicted bounding box overlaps with the ground-truth box by more than 0.5 IoU.}
        \label{tab:vg_compare}
        \scalebox{0.78}
        {
            \begin{tabular}{lccc}
            \toprule \rowcolor{mygray}
            \multicolumn{1}{c}{Method}    & RSVG  & DIOR-RSVG                  & Avg.  \\ \midrule
            QWen-VL-Chat             & 44.76 & 80.65                      & 62.71  \\
            MiniGPTv2                & 46.64 & 85.99                      & 66.32  \\ \midrule
            SkyEyeGPT                & \underline{70.50} & \textbf{88.59} & \underline{79.55}   \\ \rowcolor{cyan!50} 
            \MODELNAME               & \textbf{73.45} & \underline{88.10} &  \textbf{80.78}  \\                  \bottomrule
            \end{tabular}
        }
        \vspace{-10mm}
    \end{minipage}
\end{wrapfigure}
\subsubsection{Visual Grounding.}
The results for the visual grounding task are shown in~\cref{tab:vg_compare}, where \MODELNAME~outperforms other counterparts.
Notably, despite SkyEyeGPT being trained on a more extensive object detection dataset, \MODELNAME~demonstrates comparable performance.
This suggests that \MODELNAME~has effectively acquired fine-grained visual information throughout the training process.

\subsection{Evaluation on \BENCHMARKNAME}
We evaluate \MODELNAME~with different open-source and closed-source MLLMs on our \BENCHMARKNAME~benchmark. 
For each question within the benchmark, choices are shuffled, and the generation process is repeated four times. An answer is considered incorrect if any attempt yields an incorrect response. 
Same prompts are used for evaluation, as detailed in supplementary material.
We consider the model answers correctly if it outputs the correct choice or the answer corresponding to the correct choice. We calculate accuracy as our evaluation metric.

% The results, presented in \cref{tab:bench_result}, show that \MODELNAME~achieves the best results within all the open-source models. 
The results presented in \cref{tab:bench_result} indicate that LHRS-Bot outperforms all other open-source models.
It is worth noting that, \MODELNAME~greatly outperforms other open-source models in terms of the evaluation dimensions of spatial resolution and image modality, even though these aspects were not explicitly trained for in the models, highlighting the effectiveness of RS-specific MLLM training. Additionally, LHRS-Bot outperforms the strong closed-source model Claude and even matches the performance of GPT-4V. However, there is still significant room for improvement in MLLM capabilities within the RS domain. Examples of evaluation results are available on supplementary material.
\vspace{-2mm}

\begin{table}[t!]
\centering
\caption{Comparison results on the \BENCHMARKNAME~dataset with different open-source and closed-source MLLMs.}
\label{tab:bench_result}
\resizebox{\textwidth}{!}{%
\begin{tabular}{@{}lcccccccccccc@{}}
\toprule \rowcolor{mygray}
Method             & Identity       	& Color     & Orientation  	& Shape  	 	& Area          	& Resolution   	& Modality       	& Location  	     & Distance       & Quantity     & Reasoning	 & Avg.           \\ \midrule
LLaVA-1.5          & {42.90}  	& 30.97	& {15.38} & {43.24} 	& \textbf{44.00}	& 23.80     	& {21.74}  & {35.78} & \underline{27.27} & {21.89} & 54.35   & {37.45}    \\
MiniGPTv2          & 20.19          & 15.93		& 2.56           	& 8.11           		& 13.33          	& 14.29       	& 4.35           	& 15.20         	 & 13.64          	& 9.49          	& 36.96          & 16.88          \\
InstructBLIP       & 20.35            & 22.12	& 5.13           	& 24.32          		& 22.67          	& {38.10} & 00.00          	 & 13.73        	 & 22.85          	& 13.14        	& 30.43        & 18.87          \\
mPLUG-OWL2         & 41.48      & \textbf{45.13}	& 12.82          	& 40.54          		& {32.00} & 33.33          	& 00.00          	& 30.88          	 & 13.64          	 & \textbf{32.12}	& 54.34       & 37.00          \\
QWen-VL-Chat       & 31.38       & 32.74	& 7.69           	& 32.43         		& 29.33          	& 19.05          	& 4.35           	& 24.02          	& 21.45          	& 21.17          	& 56.52        & 28.65          \\
InternLM-XComposer & 34.38   & 30.97 	& 7.69           	& 37.84  			& 26.67          	& 00.00          	& 8.70           	& 33.33          	& {22.73}    & 18.25                & {65.22}	 & 31.09          \\ \arrayrulecolor[gray]{0.6}\cline{1-13}
\arrayrulecolor{black} 
GPT-4-Turbo          & \textbf{50.32}  	& \underline{43.36}	& 10.26 & \textbf{67.57} 	& \textbf{44.00}	& \underline{42.86}     	& \textbf{39.13}  & \underline{42.16} & {22.73} & {21.89} & \underline{71.74}   & \textbf{44.56}    \\

Claude-3-Opus           & 41.01 & 36.28 & \underline{17.95} & 64.86 		& {41.33}          	& \textbf{66.67} 	& \textbf{39.13} 	& 34.80	& \textbf{45.45}   & \underline{26.28} 	& 54.35		& 39.08 \\ \arrayrulecolor[gray]{0.3}\cline{1-13}\arrayrulecolor{black} 
\rowcolor{cyan!50}

\MODELNAME           & \underline{44.64} & {42.48} & \textbf{23.08} & \textbf{67.57} 		& 25.33          	& \underline{42.86} 	& {32.35} 	& \textbf{56.52}	& {22.73}   & 18.25 	& \textbf{73.91}		& \underline{39.38} \\ \bottomrule
\vspace{-10mm}
\end{tabular}%
}
\end{table}

\subsection{Qualitative Evaluation}
\begin{figure}[t]
  \centering
  \includegraphics[width=\linewidth]{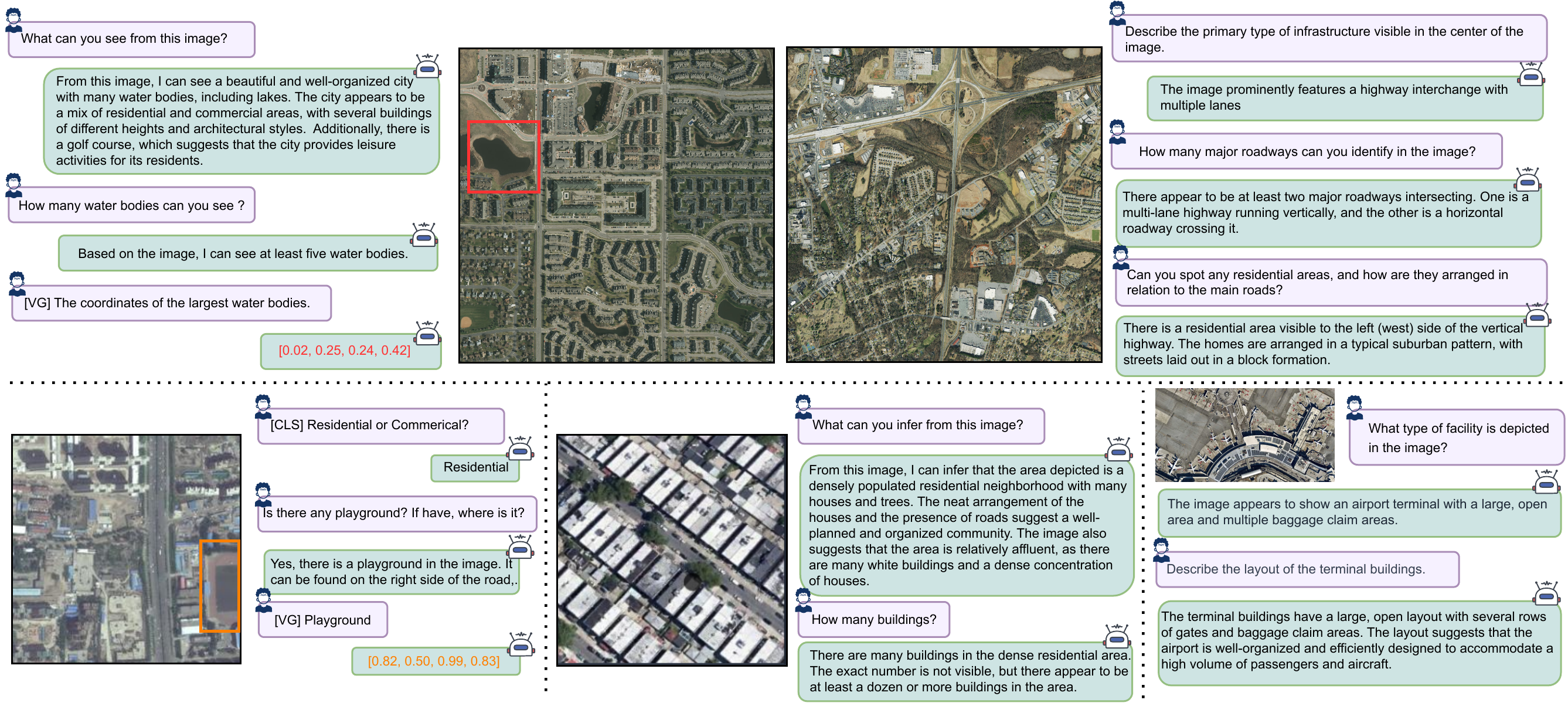}
  \caption{Qualitative examples between real users and \MODELNAME.}
  \label{fig:qualitative_exp}
  \vspace{-2mm}
\end{figure}

We conduct qualitative demonstrations to illustrate the extensive capabilities of \MODELNAME~in interpreting free-form  RS images, as showcased in \cref{fig:qualitative_exp}.
The results demonstrate that \MODELNAME~not only comprehends the overall content of an image but also exhibits a strong ability for discerning fine-grained details.
Moreover, \MODELNAME~possesses a perceptive understanding of the orientation, position, and quantity of objects within an image. 
Furthermore, \MODELNAME~is capable of making reasoned inferences from the images and acknowledges its limitations by indicating uncertainty when it cannot provide a definitive answer.
\vspace{-5mm}

\subsection{Ablation Analysis}
In this subsection, we carry out extensive ablation experiments to demonstrate the effectiveness of our dataset as well as the model design.
\vspace{-4mm}
\subsubsection{Effectiveness of \ALIGNMENTNAME~dataset.}
There are already many large-scale image-caption datasets in the field of RS~\cite{zhang2023rs5m,wang2023skyscript}.
To evaluate how different alignment datasets impact model performance, we substitute our \ALIGNMENTNAME~dataset with various image-caption datasets and repeat the entire training process. 
This process is then evaluated across three distinct tasks: classification, VQA, and visual grounding. 
The experimental setup remains consistent with~\cref{section:experi_setup}, with the exception of the dataset used in the first stage.

\begin{table}[t!]
\centering
\caption{Performance with different alignment dataset.}
\label{tab:ab_dataset}
\resizebox{\textwidth}{!}{%
\begin{tabular}{lccccccc|cc|cc}
\hline \rowcolor{mygray}
\multirow{2}{*}{}      & \multicolumn{7}{c|}{Classification}                              & \multicolumn{2}{c|}{VQA} & \multicolumn{2}{c}{Visual Grounding} \\ \cline{2-12} \rowcolor{mygray}
                              & AID   & WHU-RS19 & NWPU  & SIRI-WHU & EuroSAT & METER-ML & fMoW  & RSVQA-LR    & RSVQA-HR   & RSVG            & DIOR-RSVG          \\ \hline
RS5M                          &  86.31     &   \textbf{96.75}       &  \underline{82.80}     &    \underline{60.20}      &   \textbf{54.92}      &   \underline{64.77}       &   \underline{42.68} & 78.36 & \underline{91.65}  &     70.58        &     \underline{87.12}                  \\
SkyScript                     & \underline{88.21} & 90.26    & 78.01 & 51.43    & 43.66   & 50.02    & 31.27 & \underline{88.79}       & 91.24      & \underline{71.25}           & 85.05              \\ \rowcolor{cyan!50}
\ALIGNMENTNAME & \textbf{91.26} & \underline{93.17 }   & \textbf{83.94} & \textbf{62.66}    & \underline{51.40 }  & \textbf{69.81}    & \textbf{56.56}& \textbf{89.19}     & \textbf{92.55}     & \textbf{73.45}           & \textbf{88.10}              \\ \hline
\end{tabular}%
}
\end{table}

The results are presented in~\cref{tab:ab_dataset}.
Notably, despite possessing the smallest dataset (RS5M: 5M, SkyScript: 1.5M, \ALIGNMENTNAME~: 1.15M), \ALIGNMENTNAME~outperformes other settings.
It is  worth mentioning that both SkyScript and \ALIGNMENTNAME, which leverage RS images combined with the OSM database, achieve results comparable to or better than the even larger RS5M, which is derived from publicly available natural image datasets, such as LAION~\cite{schuhmann2022laion}.
This highlights the importance of domain-specific data collection. 
Additionally, superior performance of \ALIGNMENTNAME~over SkyScript suggests that enriched captions and a meticulous data cleaning pipeline can substantially enhance the effectiveness of MLLMs.

\subsubsection{Effectiveness of Model Design.}
We evaluate the effectiveness of designs involving multi-level visual hidden representations as well as descending query allocation strategies. Specifically, our evaluation focuses on comparing with the single level visual hidden state, projecting it through a two-layer MLP or a vision perceiver. 
We also perform comparisons between multi-level visual hidden representations, employing either identical queries for each level or incremental query allocation strategies.
Building on the concept introduced by the token merging~\cite{bolya2022token}, we explore the potential of integrating this strategy with a vision perceiver to efficiently leverage multi-level representations.
To minimize experimental costs, our training process comprises just two stages: firstly, we employ \ALIGNMENTNAME~to train the bridge layer, which aligns the vision and language domains. Subsequently, we train the model using the \SFTNAME~and multi-task instruction datasets.

\begin{table}[t!]
\centering
\caption{Performance comparison across various architectural designs.}
\label{tab:ab_model}
\centering
\scalebox{0.88}{
\begin{tabular}{@{}lccccc@{}}
\toprule \rowcolor{mygray}
Method           & Multi Level           & Number of Query    & Classification & VQA   & Visual Grounding \\ \midrule
MLP              & \xmark & -              & \underline{64.46}          & \underline{90.06} & 75.97            \\
Vision Perceiver & \xmark & 64             & 63.69          & 83.30 & 77.92            \\
Vision Perceiver & \cmark & {[}64,64,64{]} & 63.74          & 87.88 & \underline{79.30}            \\
Vision Perceiver & \cmark & {[}32,48,64{]} & 60.68          & 87.11 & 78.14            \\
Token Merging    & \cmark & 64             & 49.51          & 81.49 & 77.61            \\ \rowcolor{cyan!50}
Vision Perceiver & \cmark & {[}64,48,32{]} & \textbf{65.43}          & \textbf{90.14} & \textbf{79.90}            \\ \bottomrule
\end{tabular}%
}
\vspace{-5mm}
\end{table}

We report the average result on each task in~\cref{tab:ab_model}.
It is evident that the approach employed by \MODELNAME~achieves superior performance. 
The results further illustrate that, although the extraction of a singular visual representation performs better in tasks such as classification and VQA tasks, which depends on high-level visual semantics, it falls short in visual grounding tasks that require detailed visual information.
Furthermore, employing the same number of queries to summarise different levels of visual information does not lead to better results. In contrast, adopting a diminishing allocation strategy yields superior results. This further demonstrates the large redundancy among the deep visual tokens~\cite{bolya2022token}.
Lastly, the token merging strategy proves ineffective, possibly due to the loss of crucial information during the merging process. 

% \section{Related Work}
% \label{sec:related}
% MLLMs, specifically vision-language MLLMs, have extended the abilities of LLMs by incorporating visual alignment and visual instruction tuning~\cite{alayrac2022flamingo, li2023blip, driess2023palme,li2023llava,zhu2023minigpt,bai2023qwen}, enabling them to tackle cross-modal tasks in various domains~\cite{zhang2023xuanyuan, yang2023fingpt, wu2023bloomberggpt,yu2022legal,trautmann2022legal,thirunavukarasu2023large, singhal2023MedPalm,singhal2023MedPalm2,lin2023geogalactica, deng2023learning,li2023llava,ma2023dolphins}. 
% Concurrently, advancements in RS vision-language models, such as RSGPT~\cite{hu2023rsgpt}, GeoChat~\cite{kuckreja2023geochat}, and SkyEyeGPT~\cite{zhan2024skyeyegpt}, have achieved promising improvements in understanding RS images. 
% However, the present RS-specific MLLMs, fine-tuned using limited public task-specific datasets, neglect the potential of leveraging global RS images and vast crowdsourced databases to enhance LLMs with extensive RS visual knowledge.
% In contrast, we unleash the full potential of LLMs for RS applications by leveraging the abundant geographical information with global RS images and employing a more effective multi-level alignment strategy.
% Further details about recent related development are available in supplementary material.

\section{Conclusion}
\label{sec:conclusion}
We propose \MODELNAME, an MLLM for the RS domain. 
To unleash the potential of LLMs for RS image understanding, we curate a large-scale dataset, \ALIGNMENTNAME, for RS-specific alignment, and \SFTNAME, a multimodal instruction-following dataset to enhance \MODELNAME’s instruction-following capabilities. Additionally, we introduce \BENCHMARKNAME, a benchmark providing a systematic evaluation framework for RS MLLMs.

In comprehensive comparison experiments, \MODELNAME~demonstrates exceptional RS domain knowledge, surpassing both general-purpose and RS-specific MLLMs. Despite its impressive performance, \MODELNAME~shares limitations common to LLMs, such as susceptibility to hallucinations. 
We believe that developing higher-quality alignment and instructional datasets, along with improved training strategies, can further enhance the performance of MLLMs for interpreting RS images.

\section*{Acknowledgement}
This research was supported by the National Natural Science Foundation of China under Grant 42071297, and in part by the AI \& AI for Science Project of Nanjing University under Grant 02091480605203.
We are grateful to High Performance Computing Center of
Nanjing University for their help on GPU resources. 
We also would like to thank the anonymous reviewers
for their constructive comments.

\bibliographystyle{splncs04}
\bibliography{egbib}

\newpage
The content of each section for the appendix is as follows:
\begin{itemize}
    \item \cref{appendix:related_work} givens an in-depth review of related work.
    \item \cref{appendix: alignment data} presents explanation of the methods used to construct the \ALIGNMENTNAME~dataset.
    \item \cref{appendix: sft_dataset} outlines the steps taken to develop the \SFTNAME~dataset.
    \item \cref{appendix:multi_task_intro} details the process of converting the public remote sensing (RS) datasets into our multitask instruction dataset.
    \item \cref{appendix:test datasets} provides detailed information about the evaluation datasets.
    \item \cref{appendix:benchmark_detail} explains the evaluation dimensions of our \BENCHMARKNAME~benchmark.
    \item \cref{appendix:hyper} lists the training hyperparameters used.
    \item \cref{appendix:evaluation_prompt} includes the prompts used for evaluation on each task.
\end{itemize}
\appendix
\section{Related Work}
\label{appendix:related_work}
\subsection{Multimodal Large Language Models}
With the remarkable capabilities of LLMs in natural language processing~\cite{brown2020GPT3, chowdhery2023palm, hoffmann2022Chinchilla, zhang2022opt, workshop2022bloom, touvron2023llama} and their successful applications across various domains~\cite{zhang2023xuanyuan, yang2023fingpt, wu2023bloomberggpt,yu2022legal,trautmann2022legal,thirunavukarasu2023large, singhal2023MedPalm,singhal2023MedPalm2,lin2023geogalactica, deng2023learning}, LLMs have been extended to MLLMs by aligning visual inputs with LLMs and further empowered  by visual instruction tuning to address a serial of cross-modal tasks~\cite{alayrac2022flamingo, li2023blip, driess2023palme,li2023llava,zhu2023minigpt,bai2023qwen}.
The progression in MLLMs has catalyzed the creation of domain-specific MLLMs, encompassing areas such as biomedicine~\cite{li2023llava}, autonomous driving~\cite{ma2023dolphins}, and RS~\cite{kuckreja2023geochat, zhan2024skyeyegpt}.

\subsection{Remote Sensing Vision-Language Models}
Early RS vision-language models were designed for specific tasks, such as VQA~\cite{lobry2020rsvqa}, visual grounding~\cite{rsvg}, and scene classification~\cite{liu2023remoteclip}. 
RSGPT~\cite{hu2023rsgpt} is the initial work which exploits LLMs for RS vision-language tasks. However, it necessitates specific fine-tuning for each task, limiting its potential for holistic RS image understanding. 
Our concurrent works, GeoChat~\cite{kuckreja2023geochat} and SkyEyeGPT~\cite{zhan2024skyeyegpt} introduce RS-specific MLLMs which can also address various RS image understanding tasks and follow human instructions. 
However, these models are struggled to  adapting to existing public RS datasets, thus failing to utilize the extensive and global RS images to enrich LLMs with comprehensive RS visual knowledge.
In contrast, we unleash the full potential of LLMs for RS applications by leveraging the abundant geographical information within global available RS images and employing a more effective multi-level alignment strategy.

\subsection{Large-Scale Remote Sensing Vision-Language Datasets}
The RS5M dataset~\cite{zhang2023rs5m}, comprised of 5 million RS images, is the first large-scale RS image-text paired dataset. This dataset is curated by filtering publicly available general datasets. 
Despite the extensive scale, it falls short in providing orthorectified satellite images and offers insufficient geographical descriptions of scenes and objects. 
SkyScript~\cite{wang2023skyscript}, a dataset developed concurrent with ours, utilizes open-source geographical data to create a large-scale image-text dataset. 
However, their image captions, created using predefined rules and not thoroughly refined by a rigorous cleaning pipeline, fall short in providing high-quality and rich visual semantics. 
In this work, we develop a large-scale vision-language RS alignment dataset utilizing the semantically abundant VGI database and global RS images.
By implementing a rigorous data curation pipeline and employing LLMs to generate meaningful descriptions from feature tags, the proposed dataset is proven more informative for pretraining an MLLM.

\section{Details about~\ALIGNMENTNAME~Dataset}
\label{appendix: alignment data}
\subsection{Data Source}
\label{appendix: data source}
We have collected images that cover the most populous cities from certain countries. Due to challenges such as the scarcity of OSM data for certain cities (particularly in underdeveloped regions), the remaining images cover 9,259 cities across 129 countries. Considering the spatial resolution limitation of RS images, we only utilize the polygon features of OSM for the current version of our dataset, with the extension of polyline and point features left to our future works. Furthermore, we discard the polygon features tagged with the keys “boundary” and “barrier”, as they do not exactly describe the function of the feature. 

% \begin{figure}[h]
%   \centering
%   \includegraphics[width=\linewidth]{Figure/global1.png}
%   \caption{Global geographic scope of \ALIGNMENTNAME.}
%   \label{fig:global scope}
%   \vspace{-5mm}
% \end{figure}

\subsection{Geo-alignment and Image Processing}
\label{appendix: Geo-alignment and Image Processing}
We search the database and find out qualified features as anchors to locate the corresponding RS images. The principle is choosing features whose areas are larger than a 128×128 pixel image and have aspect ratios less than 4. This is to ensure that the chosen features are adequately large and do not exhibit excessive slenderness. For the RS images, we resize any image larger than 768×768 pixels down to 768×768. Subsequently, the images undergo a deduplication process with fastdup \footnote{\url{https://github.com/visual-layer/fastdup?tab=readme-ov-file}} and a pruning step, which involves removing images that predominantly feature vast ocean areas or are obscured by clouds. The latter process is facilitated by employing a trained classification network. These postprocessing procedures reduce the samples from 7M to 4M. Based on the geographical extent of each image, the corresponding features are selected out whose spatial area is smaller than 1/64 of the image’s size. The spatial areas of the features and images are calculated in Web Mercator projection.

\subsection{Filtering Keys}
\label{appendix: Filtering Keys}
\begin{figure}[t]
  \centering
  \fbox{\includegraphics[width=\textwidth]{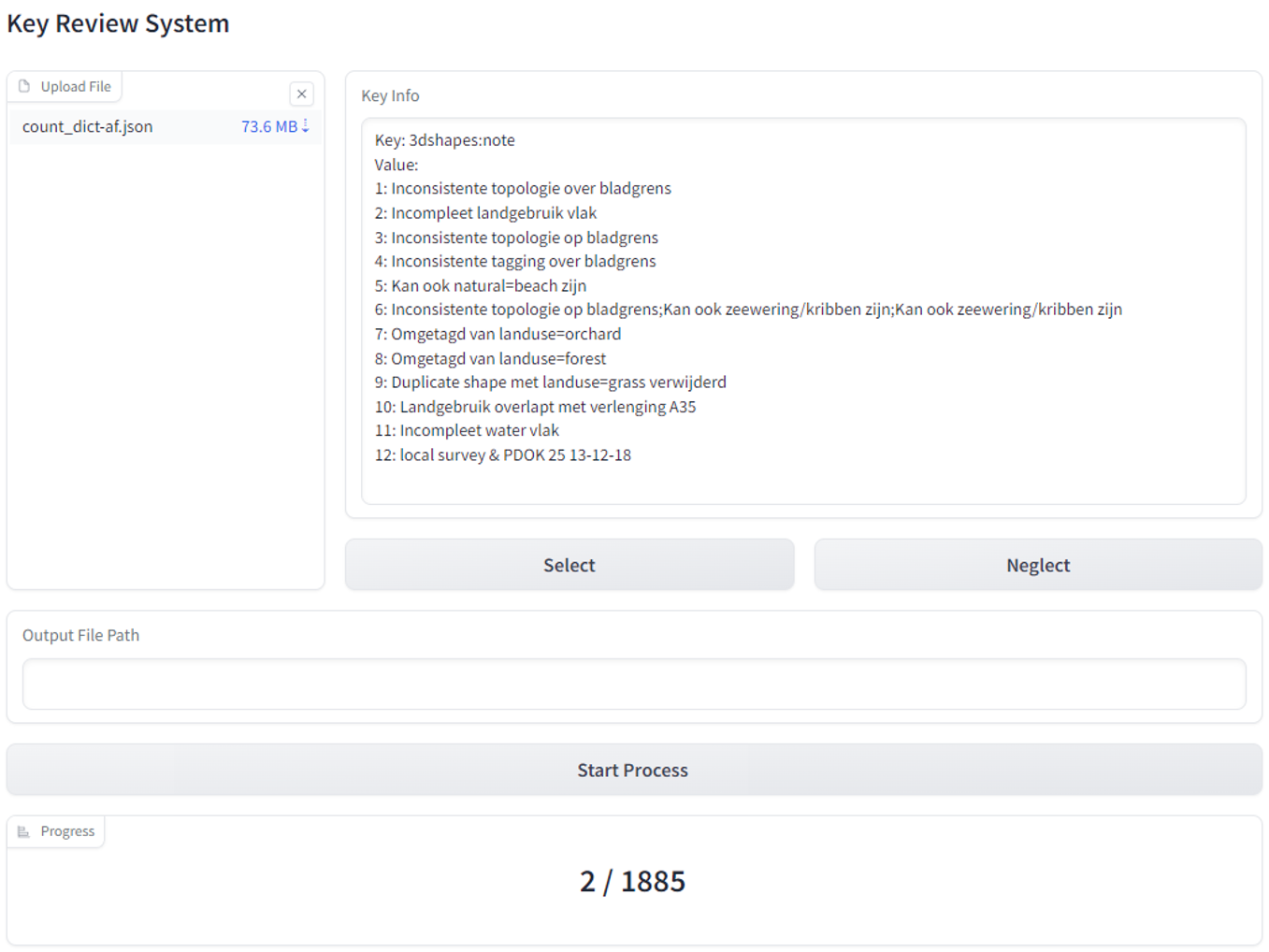}}
  \caption{Interface for manual filtering the OSM keys.}
  \label{fig:interface}
\end{figure}

We use a method combining initial automatic filtering and subsequent manual filtering. In the automatic filtering phase, we eliminate keys that meet any of these criteria: the key contains the string "name" or "addr", all values under the key have no characters, or the variety of values is fewer than 3. Through this approach, the number of keys is reduced to 1,885 from the original 10,244. In the manual filtering stage, three experts individually determine if the key is useful for providing semantic information, according to the corresponding values (via an annotation interface depicted in \cref{fig:interface}) and its description on OSM wiki. A key is preserved only if there is unanimous agreement among all three experts regarding its retention. This process further narrows the count to 186 keys, which are presented in \cref{tab:keys}. Note that this is the only step involving human discrimination throughout the pipeline, which takes only around 2 hours for each of the three experts.

\setlength{\fboxrule}{0.4mm}

\subsection{Semantic Balancing}
\label{appendix: Semantic balancing}
Due to the fact that an image corresponds to multiple geographical features, and each geographical feature corresponds to multiple attributes, we remove the duplicate key-value pairs within each image and then calculate the counts of all key-value pairs across the entire dataset. Based on the key-value statistics, we first remove the images whose key-value pairs are all larger than the magic number t = 20k (the threshold used to limit the count of key-value pairs within the whole dataset). After that, we select out images from the removed images using the independent sampling method~\cite{xu2023demystifying} for supplement to the remained images in the above step. This method results in a more balanced dataset.

\subsection{Details about Generating Captions}
\label{appendix: Prompts for Generating Captions}
To generate captions for each image in the \ALIGNMENTNAME~dataset, we utilize Vicuna-v1.5-13B, incorporating the vLLM~\cite{kwon2023efficient} and fast-chat~\cite{zheng2024judging} libraries. We explore various settings for the generation sampling parameters, adjusting the $temperature$ across $\{0.4, 0.5, 0.6, 0.7, 0.85, 0.9\}$ and $top\_p$ across $\{0.8, 0.85, 0.9, 0.95\}$. Ultimately, we select a $temperature$ of $0.7$ and a $top\_p$ of $0.95$ for optimal performance. The utilized in-context examples are detailed in~\cref{tab:cap_gen}.

\section{Details about~\SFTNAME~Dataset}
\label{appendix: sft_dataset}
\subsection{Instruction-Following Dataset from Public RS Caption Dataset}
\label{appendix:pub_sft}
\subsubsection{Data Filtering Procedure.}
The data filtering process for the public caption datasets RSITMD~\cite{yuan2022exploring} and NWPU~\cite{cheng2017remote} involves several steps:
\begin{itemize}
    \item Image deduplication is conducted using the fastdup tool.
    \item Each of the five captions for an image is tokenized, and captions shorter than 10 tokens are removed.
    \item We employ OpenAI's CLIP-L/14~\cite{radford2021learning} to calculate the average similarity between each of the five captions and their corresponding image. Samples with similarity scores below 15\% are filtered out from the dataset. This threshold was chosen after multiple attempts, as it is observed that similarity scores generated by CLIP were typically low and often did not align with human judgement.
\end{itemize}
\subsubsection{Instruction-Following Data Generation.}
We deploy the Vicuna-v1.5-13B~\cite{chiang2023vicuna} for generation with the same setting described in~\cref{appendix: Prompts for Generating Captions}. 
The in-context examples are shown in~\cref{tab:sft_pub_gen}.

\subsection{Instruction-Following Dataset from \ALIGNMENTNAME}
\label{appendix:osm_sft}
The in-context examples used for generating instruction data for visual reasoning, along with detailed image descriptions and conversations, are displayed in~\cref{tab:sft_osm_reasoning},~\cref{tab:sft_osm_detailed}, and~\cref{tab:sft_osm_conv}, respectively.

\section{Multi-task Instruction Dataset}
\label{appendix:multi_task_intro}
We build a multi-task instruction dataset by collecting various public RS datasets and creating manual instruction templates.
All the data in this multi-task instruction dataset are sampled exclusively from the training sets of each public dataset. 
The contents of the multi-task instruction dataset are detailed in~\cref{tab:multi_task_sft}.
Furthermore, different instruction templates are developed for the classification and visual grounding tasks, and each data is converted into instruction data by randomly selecting one of the templates. The instruction templates used are shown in~\cref{tab:multi_task_template}.

\section{Details about Test Datasets}
\label{appendix:test datasets}
We detail the dataset information for each task in~\cref{tab:test datasets}.

\section{Details about~\BENCHMARKNAME~Dataset}
\label{appendix:benchmark_detail}

\subsection{Evaluation Dimensions}
\label{appendix:evaluation dimensions}
The hierarchical ability taxonomies are comprised of 5 top-level dimesntions with 11 fine-grained sub-dimensions. Note that each question-answer pair may encompass multiple sub-dimensions~\cite{yu2023mmvet}. Detailed introduction of each ability dimensions is outlined below.
\begin{itemize}[label={$\bullet$}]
  \item \textbf{Recognition:} This dimension focuses on the model’s ability to recognize instance identity (geographical objects or scene information), and their attributes, including color, orientation, shape, and area.
  \item \textbf{Quantity:} This dimension assesses the model’s capacity to count specific objects, often combined with a prerequisite understanding of those objects (Recognition).
  \item \textbf{Imagery:} This dimension centers on the image itself, specifically recognizing its resolution and modality. In the RS domain, images can encompass various modalities, including optical, panchromatic, synthetic aperture radar (SAR), and thermal infrared images. However, as RS MLLMs are still in their early stages of development, \BENCHMARKNAME~currently only includes multispectral and panchromatic images.
  \item \textbf{Spatial-awareness:} This dimension evaluates the MLLM’s ability to comprehend the locations (and spatial relationships between objects), as well as identify the distance between objects.
  \item \textbf{Reasoning:} This dimension assesses the model’s proficiency in accurately comprehending the image and utilizing its domain-specific knowledge to provide correct answers to questions.
\end{itemize}

% \subsection{Prompt for Evaluation}
% \label{benchmark prompt}

\section{Hyperparamters}
\label{appendix:hyper}
In all our experiments, we utilized 8 NVIDIA V100-32G GPU. However, due to the V100's lack of support for bf16, we initially trained using the AdamW optimizer and fp16. This approach, however, led to considerable instability in the first stage of training. We experimented with various learning rates, specifically \{$2\times 10^{-6}$, $5\times 10^{-6}$, $8\times 10^{-6}$, $2\times 10^{-5}$, $5\times 10^{-5}$, $8\times 10^{-5}$, $2\times 10^{-4}$, $1\times 10^{-3}$\}, but observed that training became unstable at higher learning rates, while lower learning rates resulted in suboptimal model performance. We also implemented gradient clipping and experimented with different gradient norms, specifically \{0.1, 0.3, 0.6, 1.0, 5.0\}. Ultimately, for improved performance and stability in the first stage, we selected the Adan optimizer~\cite{xie2022adan}. Concurrently, we enabled DeepSpeed-Zero2~\cite{rajbhandari2020zero} in all three training stages. Details of other hyperparameter settings can be found in Table~\ref{tab:hyper}.

\section{Evaluation Prompt for Each Task}
\label{appendix:evaluation_prompt}
In tasks related to visual grounding and VQA, we prompt each model using descriptions of the target object and questions, respectively. For classification tasks and evaluations on \BENCHMARKNAME, the same prompts are employed across all models, with the specific prompts presented in~\cref{tab:prompt_4_each_task}.

\newcolumntype{H}{>{\setbox0=\hbox\bgroup}c<{\egroup}@{}}

\begin{table}[ht!]
\caption{Detailed statistics of multi-task instruction dataset.}
\centering
\label{tab:multi_task_sft}
\begin{tabular}{@{}lcc@{}}
\toprule \rowcolor{mygray}
Dataset   & Task Type              & Samples \\ \midrule
RSVQA-HR~\cite{lobry2020rsvqa}  & VQA                    & 10,000  \\
RSVQA-LR~\cite{lobry2020rsvqa}  & VQA                    & 500     \\
UCM~\cite{qu2016UCM}       & Classification, Caption & 2,519   \\
RSVG~\cite{sun2022VG}      & Visual Grounding       & 2,428   \\
DIOR-RSVG~\cite{rsvg} & Visual Grounding       & 14,030  \\
NWPU~\cite{cheng2017remote}      & Classification         & 4,941   \\
METER-ML~\cite{zhu2022meter}   & Classification         & 1,400   \\
RSITMD~\cite{yuan2022exploring}    & Classification         & 504     \\
fMoW~\cite{christie2018functional}     & Classification         & 5,000   \\
RSICD~\cite{lu2017rsicd}     & Caption                & 1,000   \\ \midrule
Total     &                        & 42,322  \\ \bottomrule
\end{tabular}
\end{table}

\begin{table}[ht!]
\caption{Detailed introduction of the test datasets.}
\centering
\label{tab:test datasets}
\begin{tabular}{@{}lcccc}
\toprule \rowcolor{mygray}
Dataset   & Task Type           & Image size & Resolution (m)       & Categories    \\ \midrule
AID~\cite{xia2017aid}       & Classification      & 600×600    & 0.5 -- 8       & 30    \\         
WHU-RS19~\cite{Dai2011WHURS19}  & Classification      & 600×600    & -- 0.5     & 19  \\
NWPU~\cite{cheng2017remote}      & Classification      & 256×256    & 0.2 -- 30    & 45          \\
SIRI-WHU~\cite{zhu2016SIRI-WHU}  & Classification      & 200×200    & 2        & 12      \\
EuroSAT~\cite{helber2019eurosat} & Classification      & 64×64     & 10         & 10    \\
METER-ML~\cite{zhu2022meter}  & Classification      & 720×720    & 1            & 6     \\
fMoW~\cite{christie2018functional}  & Classification  & Varies     & 0.31 -- 1.6   &  63   \\
RSVQA-LR~\cite{lobry2020rsvqa}  & VQA                 & 256×256    & 10         & --    \\
RSVQA-HR~\cite{lobry2020rsvqa}  & VQA                 & 512×512    & 0.15      & --   \\
RSVG~\cite{sun2022VG}      & Visual Grounding    & 1024×1024  & Varies        & --        \\
DIOR-RSVG~\cite{rsvg} & Visual Grounding    & 800×800    & 0.5 -- 30        & 20  \\      \bottomrule
\end{tabular}
\end{table}

\begin{table}[ht!]
\caption{Hyperparameter settings.}
\centering
\label{tab:hyper}
\begin{tabular}{@{}lccc@{}}
\toprule \rowcolor{mygray}
                     & Stage1  & Stage2     & Stage3      \\ \midrule
Optimizer            & Adan    & AdamW      & AdamW       \\
Learning Rate        & 0.0002  & 0.0002     & 0.0001      \\
Batch Size           & 8       & 4          & 4           \\
Accumulation Step(s) & \multicolumn{3}{c}{1}              \\
Weight Decay         & \multicolumn{3}{c}{0.0}            \\
$\beta_1$             & 0.98    & \multicolumn{2}{c}{0.9}  \\
$\beta_2$             & 0.92    & \multicolumn{2}{c}{0.95} \\
$\beta_3$             & 0.99    & \multicolumn{2}{c}{-}    \\
Epoch(s)/Step(s)     & 1 Epoch & 1 Epoch    & 1200 Steps  \\
Gradient Norm        & 0.3     & 1.0        & 1.0         \\
Scheduler            & \multicolumn{3}{c}{Cosine}         \\
Warmup Steps         & 500     & 100        & 40          \\ \bottomrule
\end{tabular}
\end{table}

\begin{table*}[h!t]
    \centering
    \begin{minipage}{0.99\columnwidth}
    \vspace{0mm}    
    \centering
    \caption{Selected keys for OSM features.}
    \vspace{-5mm}
    \label{tab:keys}
    \begin{tcolorbox} 
        \centering
        \small
        \hspace{-6mm}
        \begin{tabular}{p{0.99\columnwidth}}
            \begin{minipage}{0.99\columnwidth}\vspace{0mm}

                \VarSty{Manual selected keys:}
                
                [ 'landuse', 'landcover', 'natural', 'amenity', 'water', 'industrial', 'product', 'leisure', 'tourism', 'shop', 'healthcare', 'aeroway', 'surface', 'aerodrome', 'highway', 'building', 'sport', 'denomination', 'man\_made', 'waterway', 'power', 'military', 'parking', 'historic', 'residential', 'trees', 'place', 'plant:source', 'public\_transport', 'leaf\_cycle', 'produce', 'mineral', 'social\_facility', 'castle\_type', 'wetland', 'ruins', 'archaeological\_site', 'water\_source', 'type', 'crop', 'historic:civilization', 'material', 'area:highway', 'education', 'construction', 'garden:type', 'museum', 'resort', 'leaf\_type', 'seamark:type', 'service', 'military\_service', 'beach', 'craft', 'generator:source', 'resource', 'cargo', 'golf', 'orchard', 'railway', 'grassland', 'depot', 'port', 'attraction', 'content', 'roof:shape', 'meadow', 'wood', 'grass', 'zoo', 'construction:power', 'generator:method', 'plant', 'aquaculture', 'reservoir\_type', 'sport\_1', 'school', 'trade', 'basin', 'piste:type', 'club', 'monastery:type', 'landform', 'athletics', 'wholesale', 'building:use', 'pumping\_station', 'utility', 'police', 'company', 'playground', 'footway', 'building:colour', 'construction:amenity', 'animal\_shelter', 'cemetery', 'tomb', 'landfill', 'construction:leisure', 'building:material', 'landfill:waste', 'construction:landuse', 'farmyard', 'shelter\_type', 'camp\_type', 'animal\_keeping', 'building:part', 'waste', 'animal\_breeding', 'construction:man\_made', 'commercial', 'research', 'species', 'seamark:small\_craft\_facility:category', 'glacier:type', 'hazard', 'geological', 'reservoir', 'historic:landuse', 'station', 'recreation\_ground', 'construction:railway', 'boules', 'roof:material', 'park', 'amenity\_1', 'animal', 'usage', 'contamination', 'seamark:shoreline\_construction:category', 'reef', 'farmland', 'nature', 'storage', 'livestock', 'pasture', 'allotments', 'building:condition', 'logistics', 'forest', 'building:roof', 'place\_of\_worship', 'rock', 'factory', 'scrub', 'construction:industrial', 'seaway', 'construction:sport', 'park:type', 'vehicle\_depot', 'dock', 'bunker\_type', 'toponym', 'industry', 'historic:leisure', 'sand', 'earth\_bank', 'pond\_use', 'tree', 'retail', 'bay', 'works', 'forest\_cover', 'historic:amenity', 'construction:residential', 'sports\_centre', 'residence', 'paving\_stones:material', 'seamark:sea\_area:category', 'quarry', 'vegetation', 'construction:highway', 'pitch', 'warehouse', 'construction:shop', 'bridge:type', 'historic:aeroway', 'bare\_rock', 'transport', 'bridge:movable', 'construction:tourism', 'building:roof:colour', 'amenity\_2', 'amenity\_3', 'construction:type', 'heath' ]

            \end{minipage}
    \end{tabular}
    \end{tcolorbox}
    \end{minipage}
\end{table*}
\vspace{-7mm}

\begin{table*}[!ht]
    \centering
    \begin{minipage}{0.99\columnwidth}
    \vspace{0mm}    
    \centering
    \caption{In-context examples for~\ALIGNMENTNAME~caption generation.}
    \vspace{-5mm}
    \label{tab:cap_gen}
    \begin{tcolorbox} 
        \centering
        \small
        \hspace{-6mm}
        \begin{tabular}{p{0.99\columnwidth}}

            \begin{minipage}{0.99\columnwidth}\vspace{0mm}

                \VarSty{System Message:}
                
                As an AI visual assistant, your task is to create a caption for a single image presented to you. This image is accompanied by several key-value tags, each providing specific details about it. Your goal is to synthesize these details into a clear, concise description that encompasses all the features mentioned in the key-value tags. The description should be diverse, reflecting the variety of attributes and elements present in the image. Finish this task as if you are directly observing the image yourself, and craft a caption that naturally integrates all the provided information. Remember, your description should not explicitly reference the key-value pairs; instead, it should read as a seamless and direct observation of the image, capturing its essence in a fluent and engaging manner.

                \vspace{2mm}
                \VarSty{In-Context Example:}
                
                \VarSty{User:}

                There are four tags contained in this image. Their keys and values are listed below:\textbackslash n1. Key: landuse, Value: residential\textbackslash n2.Key: natural, Value: water; Key: Water, Value: River\textbackslash n3.Key: landuse, Value: commerical\textbackslash n4.Key: landuse, Value: farmland.

                \VarSty{Assistant:}

                The image presents a varied landscape featuring a residential area, a river representing natural water bodies, commercial zones with business or retail establishments, and farmland used for agricultural purposes.

                \VarSty{User:}

                There are four features in the images. Their keys and values are listed below:\textbackslash n1.Key: landuse, Value: residential\textbackslash n2.Key:landuse, Value: farmland\textbackslash n3.Key: landuse, Value: industrial\textbackslash n4.Key:industrial, Value: factory; Key: landuse, Value: industrial; Key: product, Value: bricks.

                \VarSty{Assistant:}

                The image shows a varied landscape with a residential area, farmland, and an industrial zone featuring a brick factory, illustrating a blend of community, agriculture, and industry.        
            \end{minipage}
        \end{tabular}
    \end{tcolorbox}
    \end{minipage}
    \vspace{-10mm}
\end{table*}

\begin{table*}[!ht]
    \centering
    \begin{minipage}{0.99\columnwidth}
    \vspace{0mm}    
    \centering
    \caption{In-context examples for caption generation from the public caption datasets.}
    \vspace{-5mm}
    \label{tab:sft_pub_gen}
    \begin{tcolorbox} 
        \centering
        \small
        \hspace{-6mm}
        \begin{tabular}{p{0.99\columnwidth}}

            \begin{minipage}{0.99\columnwidth}\vspace{0mm}

                \VarSty{System Message:}
                
                    You are an AI visual assistant, and you are seeing a single image. What you see are provided with ﬁve sentences, describing the same image you are looking at. Answer all questions as you are seeing the image.
                    
                    Design a conversation between you and a person asking about this photo. The answers should be in a tone that a visual AI assistant is seeing the image and answering the question. Ask diverse questions and give corresponding answers.
                    
                    Include questions asking about the visual content of the image, including the object types, object actions, object locations, relative positions between objects, etc. Only include questions that have deﬁnite answers:
                    
                    (1) one can see the content in the image that the question asks about and can answer conﬁdently;
                    
                    (2) one can determine conﬁdently from the image that it is not in the image. Do not ask any question that cannot be answered conﬁdently.

                \vspace{2mm}
                \VarSty{In-Context Example:}
                
                \VarSty{User:}

                There is a baseball field beside the green amusement park around the red track.
                
                A green baseball field adjacent to the playground and Red Square.
                
                There is a long path in the field next to the red playground.
                
                The green playground around the red runway is a baseball field.
                
                The green baseball field is adjacent to the playground and the red playground.

                \VarSty{Assistant:}
                
                \textbf{Questions:}

                Please give a detailed description about this image.

                \textbf{Answer:}

                The image shows large, open field with a baseball diamond in the center. The field i surrounded by trees, and there is a dirt path leading to the diamond. The field is likely a sports field, possibly used for baseball or other outdoor sports. The presence of the baseball diamond suggests that it is a baseball field, but it could also be used for other sports like soccer or football.

                \textbf{Questions:}
                
                What is the color of the main object in this image?

                \textbf{Answer:}

                The amusement park and baseball field is green, whereas the playground is depicted in red.

                \textbf{Questions:}
                What does the image suggest about the function and superiority about this image?

                \textbf{Answer:}
                
                The image captures the natural beauty of the area, with the trees providing shade and the open field offering ample space for sports activities. The dirt path leading to the diamond adds a rustic touch to the scene, creating a picturesque and inviting environment for sports enthusiasts.      
            \end{minipage}
        \end{tabular}
    \end{tcolorbox}
    \vspace{-10mm}
    \end{minipage}
\end{table*}

\begin{table*}[!ht]
    \centering
    \begin{minipage}{0.99\columnwidth}
    \vspace{0mm}    
    \centering
    \caption{In-context examples for generating visual reasoning samples using GPT-4.}
    \vspace{-5mm}
    \label{tab:sft_osm_reasoning}
    \begin{tcolorbox} 
        \centering
        \fontsize{8pt}{10pt}\selectfont
        \hspace{-6mm}
        \begin{tabular}{p{0.99\columnwidth}}
                \VarSty{System Message:}
                
                    You are an AI visual assistant that can analyze a single image. You receive one sentence, describing the same image you are observing. In addition, specific object locations and their attributes within the image are given, along with detailed coordinates. These coordinates are in the form of bounding boxes, represented as (x1, y1, x2, y2) with floating numbers ranging from 0 to 1. These values correspond to the top left x, top left y, bottom right x, and bottom right y.
                    
                    The task is to use the provided caption and bounding box information, create a plausible question about the image, and provide the answer in detail.
                    
                    Create complex questions beyond describing the scene.
                    To answer such questions, one should require first understanding the visual content, then based on the background knowledge or reasoning, either explain why the things are happening that way, or provide guides and help to user's request.  Make the question challenging by not including the visual content details in the question so that the user needs to reason about that first.
                    Also include some relations about relative position, distance among different objects.
                    
                    Instead of directly mentioning the bounding box coordinates, utilize this data to explain the scene using natural language. Include details like object counts, position of the objects, relative position between the objects and their relations.  
                    
                    When using the information from the caption and coordinates, directly explain the scene, and do not mention that the information source is the caption or the bounding box.  Always answer as if you are directly looking at the image.

                \vspace{2mm}
                \VarSty{In-Context Example:}
                
                \VarSty{User:}

                The image portrays a combination of different land uses and recreational areas, including a stadium used for basketball, a marina, a beach, a grassland area, and a parking lot with surface parking. Additionally, there is a harbor serving as a seamark and a park for leisure purposes.
                
                building: stadium,leisure:stadium,roof:shape:flat,sport:basketball -> [0.0, 0.249, 1.0, 0.757]
                
                leisure:marina,seamark:type:harbour -> [0.777, 0.0, 1.0, 0.373]
                
                natural:beach -> [0.137, 0.0, 0.792, 0.201]
                
                natural:grassland -> [0.196, 0.903, 1.0, 0.956]
                
                amenity:parking,parking:surface -> [0.561, 0.659, 0.956, 0.776]
                
                leisure: park -> [0.0, 0.0, 1.0, 0.887]

                \VarSty{Assistant:}
                
                \textbf{Questions:}

                How do the diverse land uses and recreational facilities depicted in the image cater to the varied interests and needs of the community, and what role does the layout play in the functionality and aesthetic appeal of the area?
                
                \textbf{Answer:}

                The image's diverse land uses cater to various interests with thoughtful placement: the basketball stadium, centrally located, hosts sports and events; the marina and beach, positioned to the north, offer water activities and relaxation; the grassland, towards the south, provides open space for passive leisure. The nearby parking, situated to the southeast, facilitates easy access, while the park, enveloping much of the area, ties these elements together with greenery and walking paths. This strategic layout balances activity and tranquility, effectively serving the community's needs and enhancing the area's aesthetic and functional appeal.   
        \end{tabular}
    \end{tcolorbox}
    \vspace{-2mm}
    \end{minipage}
\end{table*}
% Detail table
\begin{table*}[!ht]
    \centering
    \begin{minipage}{0.99\columnwidth}
    \vspace{0mm}    
    \centering
    \caption{In-context examples for generating detailed image descriptions using GPT-4.}
    \vspace{-5mm}
    \label{tab:sft_osm_detailed}
    \begin{tcolorbox} 
        \centering
        \fontsize{7.6pt}{10pt}\selectfont
        \hspace{-6mm}
        \begin{tabular}{p{0.99\columnwidth}}
                \VarSty{System Message:}
                
                    You are an AI visual assistant with the capability to analyze complex images. For this task, you will be given descriptions and precise object location data of a single image. The image will be described through five distinct sentences, each providing a unique perspective or detail about the same scene. Accompanying these descriptions are specific object locations within the image, denoted by bounding box coordinates. These coordinates are formatted as (x1, y1, x2, y2), where each value is a floating number between 0 and 1, representing the proportions of the image's dimensions. The coordinates x1 and y1 refer to the top left corner of the bounding box, while x2 and y2 refer to the bottom right corner.

                    Your task is to summarize the information from the sentences and the bounding box data to create a rich, detailed description of the scene. In your description, focus on the layout, the spatial relationships between objects, and any notable details that the combination of textual descriptions and coordinate data reveals. Remember, the goal is to paint a vivid picture of the scene as if you are observing it firsthand. Avoid explicitly mentioning the source of your information, such as the specific sentences or the bounding box coordinates. Instead, integrate this information seamlessly into your natural language description, emphasizing the number of objects, their exact positions, and how they relate to one another in the space.

                    Answer as if you're guiding someone through the image, pointing out each element and its significance in the overall scene. Your description should enable someone who cannot see the image to visualize it as accurately and completely as possible.

                \vspace{2mm}
                \VarSty{In-Context Example:}
                
                \VarSty{User:}

                The image portrays a combination of different land uses and recreational areas, including a stadium used for basketball, a marina, a beach, a grassland area, and a parking lot with surface parking. Additionally, there is a harbor serving as a seamark and a park for leisure purposes.
                
                building: stadium,leisure:stadium,roof:shape:flat,sport:basketball -> [0.0, 0.249, 1.0, 0.757]
                
                leisure:marina,seamark:type:harbour -> [0.777, 0.0, 1.0, 0.373]
                
                natural:beach -> [0.137, 0.0, 0.792, 0.201]
                
                natural:grassland -> [0.196, 0.903, 1.0, 0.956]
                
                amenity:parking,parking:surface -> [0.561, 0.659, 0.956, 0.776]
                
                leisure: park -> [0.0, 0.0, 1.0, 0.887]

                \VarSty{Assistant:}
                
                \textbf{Questions:}

                What are the key features and elements depicted in the urban landscape captured in the image?
                
                \textbf{Answer:}

                The image depicts a vibrant urban landscape, harmoniously integrating various recreational and natural elements. Dominating the scene is a large basketball stadium with a flat roof, stretching from the top to nearly the bottom and spanning the entire width, signifying its prominence. Adjacent to it, in the top right corner, is a marina, hinting at a coastal setting. Just below the marina, a beach area extends almost the full width of the image, highlighting its importance as a recreational spot.

                At the very bottom, a narrow strip of grassland provides a natural contrast to the urban environment. Centrally located in the lower middle part is a surface parking lot, strategically placed for easy access to the surrounding facilities. Overlaying much of the image is a leisure park, suggesting a large, open area designed for relaxation and outdoor activities.

                Together, these elements create a dynamic and multifunctional urban space, blending sports, leisure, and nature.
        \end{tabular}
    \end{tcolorbox}
    \vspace{-3mm}
    \end{minipage}
\end{table*}
% Conv table
\begin{table*}[!ht]
    \centering
    \begin{minipage}{0.99\columnwidth}
    \vspace{0mm}    
    \centering
    \caption{In-context examples for generating multimodal conversations using GPT-4.}
    \vspace{-5mm}
    \label{tab:sft_osm_conv}
    \begin{tcolorbox} 
        \centering
        \fontsize{8pt}{10pt}\selectfont
        \hspace{-6mm}
        \begin{tabular}{p{0.99\columnwidth}}
                \VarSty{System Message:}
                You are an AI visual assistant that can analyze a single image. You receive one sentence, describing the same image you are observing. In addition, specific object locations and their attributes within the image are given, along with detailed coordinates. These coordinates are in the form of bounding boxes, represented as (x1, y1, x2, y2) with floating numbers ranging from 0 to 1. These values correspond to the top left x, top left y, bottom right x, and bottom right y.

                Your task is to engage in a conversation where you design questions and provide answers about the visual content of the image. The questions should focus on object count, position, relative distance, and other visual aspects, while the answers should reflect direct observation as if you are viewing the image. Create questions that delve into the visual details of the image by considering the types of objects present, their actions, and their spatial relationships.
                
                Provide answers as if you're observing the image firsthand. For straightforward questions, give concise and accurate responses. For more complex inquiries, offer detailed explanations or step-by-step reasoning to justify your answers.
                
                \vspace{2mm}
                \VarSty{In-Context Example:}
                
                \VarSty{User:}

                The image shows a mix of land uses, including a bus depot, industrial areas, construction sites, brownfield land with parking aisles, meadows, retail spaces, and a bus station as a public transport hub. The natural feature in the scene is scrub, which is a type of vegetation.
                
		      depot:bus,landuse:depot -> [0.169, 0.0, 0.832, 1.0]
        
	       	landuse:industrial -> [0.0, 0.717, 0.254, 0.86]
         
		      natural:scrub -> [0.0, 0.504, 0.314, 0.86]
        
                landuse:construction -> [0.276, 0.02, 0.479, 0.254]
                
		      landuse:industrial -> [0.751, 0.1, 1.0, 0.67]
        
                landuse:meadow -> [0.661, 0.864, 1.0, 1.0]
                
                landuse:retail -> [0.687, 0.629, 1.0, 0.916]
                
                \VarSty{Assistant:}
                
                \textbf{Question:}
                How many distinct industrial areas are there in the image, and where are they located relative to the entire scene?

                \textbf{Answer:}
                There are two distinct industrial areas. One is towards the bottom left corner, extending up and to the right but not reaching the image's center. The other is towards the right side, stretching from near the bottom up to about two-thirds of the height.

                \textbf{Question:}
                What is the relative position of the bus depot to the construction site?

                \textbf{Answer:}
                The bus depot is centrally located, running vertically almost across the entire image height. The construction site is situated towards the upper left, overlapping with the lower part of the bus depot. It's closer to the image's top edge than the bottom.

                \textbf{Question:}
                Is there any natural vegetation, and if so, where is it located in relation to the industrial areas?

                \textbf{Answer:}
                Yes, natural vegetation identified as scrub is present towards the left side, extending from near the middle to the bottom. It overlaps with the first industrial area, indicating they are adjacent or possibly intermingled.

               \textbf{Question:}
                Can you describe the distribution of land for retail and meadows?

                \textbf{Answer:}
                The retail space is in the upper right quarter, not quite reaching the top or rightmost edge. The meadow is also in the upper right but is situated higher and further to the right than the retail space, with a clear separation between the two areas indicating distinct zones for commerce and natural space.
        \end{tabular}
    \end{tcolorbox}
    \vspace{-3mm}
    \end{minipage}
\end{table*}

\begin{table*}[!ht]
    \centering
    \begin{minipage}{0.99\columnwidth}
    \vspace{0mm}    
    \centering
    \caption{Templates for constructing multi-task instruction dataset.}
    \vspace{-5mm}
    \label{tab:multi_task_template}
    \begin{tcolorbox} 
        \centering
        \fontsize{8pt}{10pt}\selectfont
        \hspace{-6mm}
        \begin{tabular}{p{0.99\columnwidth}}
                \VarSty{Classification:}
                
                Create a class label for the image from the following categories: \{\}.
                
                Classify the image into the following categories: \{\}.
                
                You have been provided with the following categories: \{\}.
                
                Choose the best categories to describe the image from: \{\}.
                
                Please suggest a suitable class name based on the provided class type for this image: \{\}.
                
                Assign a class label to the image from the available categories: \{\}.
                
                Categorize the image into the subsequent groups: \{\}.
                
                You are given the following categories to classify the image: \{\}.
                
                Select the most fitting categories that define the image from: \{\}.
                
                Please suggest a suitable class name based on the provided class type for this image: \{\}.
                
                Assign a class label to the image from the available categories: \{\}.
                
                Categorize the image into the subsequent groups: \{\}.
                
                You are given the following categories to classify the image: \{\}.
                
                \vspace{2mm}
                \VarSty{Visual Grounding:}
                
                Please output the coordinate of the following object: \{\}.
                
                Kindly provide the position for the object mentioned below: \{\}.
                
                I request the location for the object outlined here: \{\}.
                
                Could you please share the placement of the object depicted: \{\}.
                
                The details for the following object are needed: \{\}.
                
                Please provide the dimensions corresponding to the following object: \{\}.
                
                I require the spatial information of the object below: \{\}.
                
                Kindly share the positional data for the object shown: \{\}.
                
                The object's placement information is necessary: \{\}.
                
                Could you output the location of the object below: \{\}.
                
                I need the position of the object mentioned: \{\}.
                
                Share the placement of the object outlined here: \{\}.
                
                Please provide the spatial data for the object depicted: \{\}.
                
                I'm looking for details about the following object: \{\}.
                
                Kindly provide the dimensions for the object shown: \{\}.
                
                It's important to have the spatial coordinates for the object: \{\}.
        \end{tabular}
    \end{tcolorbox}
    \vspace{-3mm}
    \end{minipage}
\end{table*}

\begin{table*}[!ht]
    \centering
    \begin{minipage}{0.99\columnwidth}
    \vspace{0mm}    
    \centering
    \caption{Prompts for evaluation on each task.}
    \vspace{-5mm}
    \label{tab:prompt_4_each_task}
    \begin{tcolorbox} 
        \centering
        \fontsize{8pt}{10pt}\selectfont
        \hspace{-6mm}
        \begin{tabular}{p{0.99\columnwidth}}
                \VarSty{Classification:}

                Choose the best categories describe the image from: \{\}.
                
                \vspace{2mm}
                \VarSty{\BENCHMARKNAME:}

                Please answer the question based on the given choices:

                Question: \{\}

                Choice: \{\}

                Answer:

        \end{tabular}
    \end{tcolorbox}
    \vspace{-3mm}
    \end{minipage}
\end{table*}

\begin{figure}[ht!]
  \centering
  \includegraphics[width=0.95\textwidth]{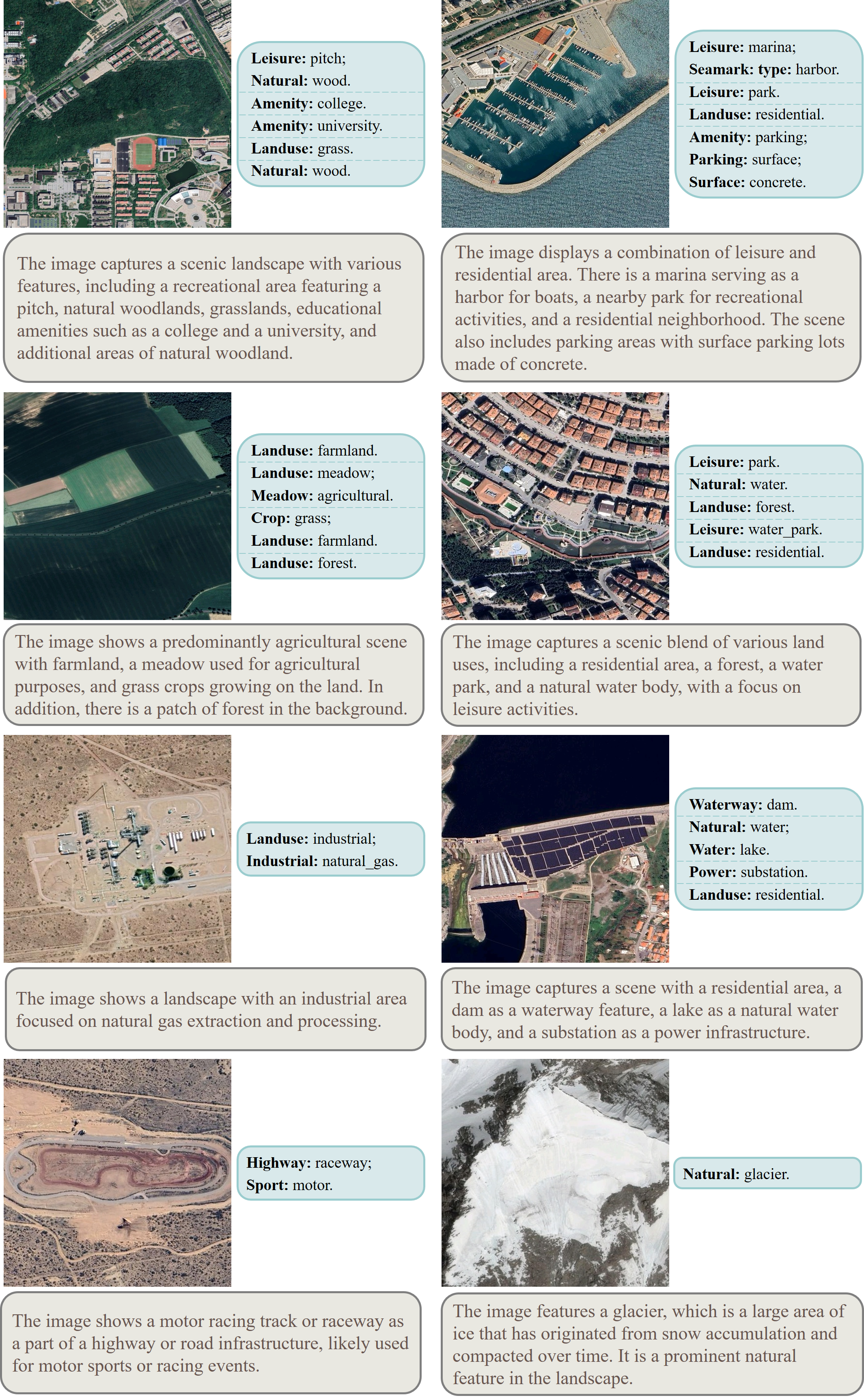}
  \caption{Examples of the \ALIGNMENTNAME~dataset.}
  \label{fig:alignment}
\end{figure}

\begin{figure}[t!]
  \centering
  \includegraphics[width=0.95\textwidth]{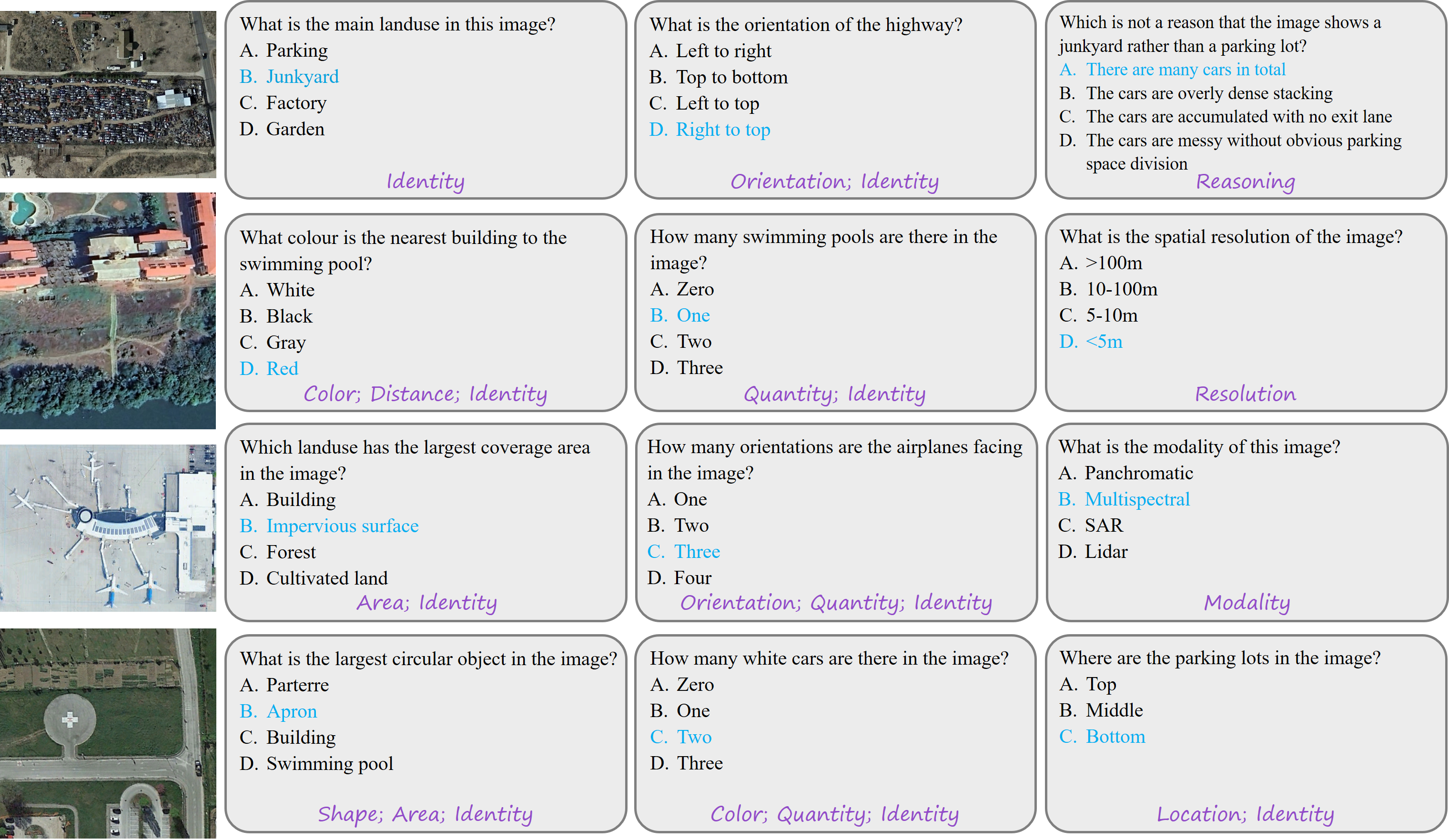}
  \caption{Examples of the \BENCHMARKNAME~benchmark.}
  \label{fig:benchmark}
\end{figure}

\begin{figure}[t!]
  \centering
  \includegraphics[width=0.95\textwidth]{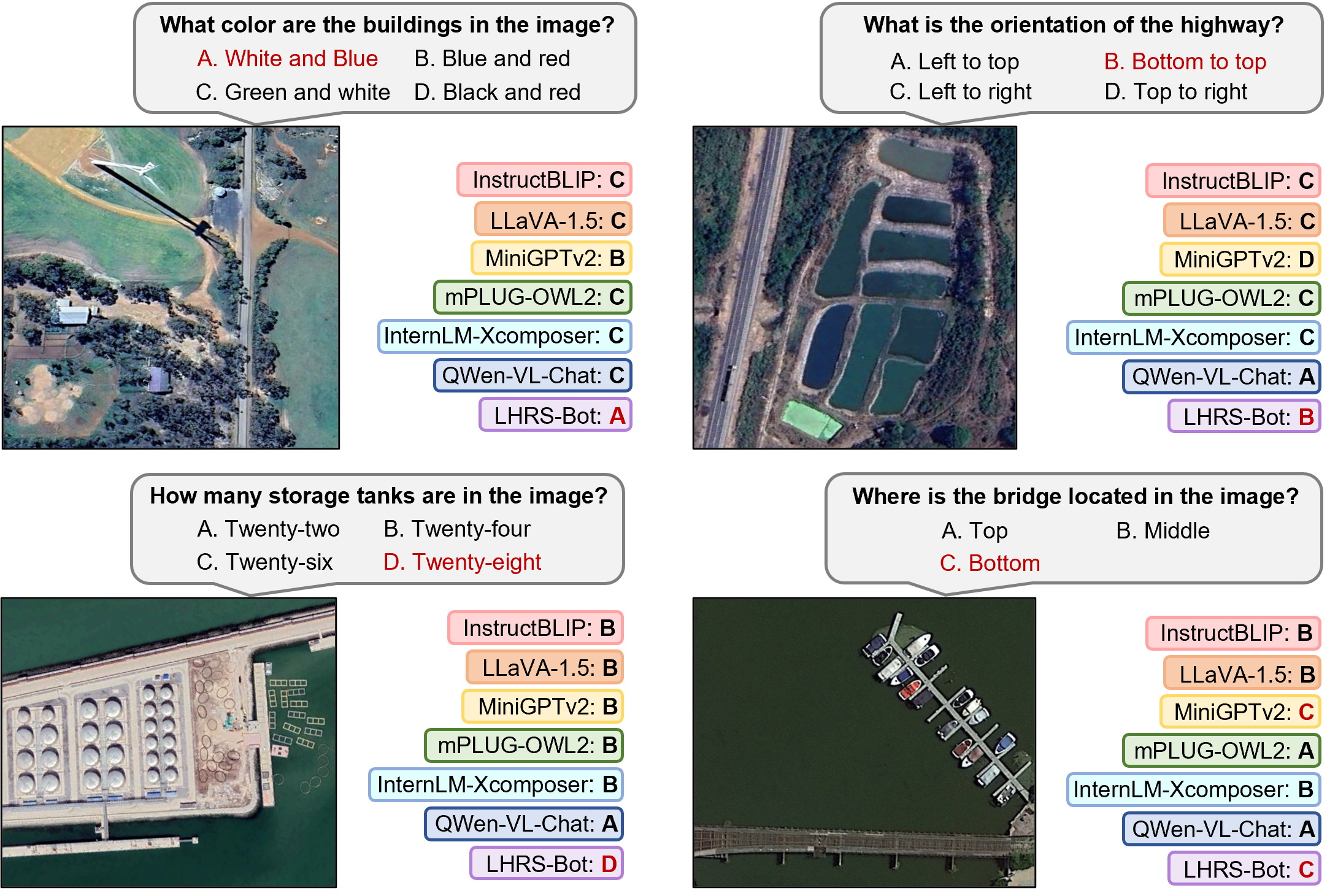}
  \caption{Evaluation examples on~\BENCHMARKNAME~benchmark.}
  \label{fig:benchmark_result}
\end{figure}

\end{document}